\documentclass{article} 
\usepackage{preprint,times}

\usepackage{amsmath,amsfonts,bm}

\def\eqref#1{equation~\ref{#1}}

\def\1{\bm{1}}

\DeclareMathAlphabet{\mathsfit}{\encodingdefault}{\sfdefault}{m}{sl}
\SetMathAlphabet{\mathsfit}{bold}{\encodingdefault}{\sfdefault}{bx}{n}

\usepackage{hyperref}
\usepackage{url}
\usepackage{amsmath}
\usepackage{amssymb}
\usepackage{booktabs}
\usepackage{multirow}
\usepackage{graphicx}
\usepackage[table]{xcolor}
\usepackage{caption}
\usepackage{array}
\usepackage{algorithm}
\usepackage{algpseudocode}
\usepackage{pifont}
\usepackage{graphicx}
\usepackage{subcaption}
\usepackage{wrapfig}
\definecolor{verylightgray}{gray}{0.89} 
\usepackage[table]{xcolor}
\usepackage{multirow}
\usepackage{amssymb}
\usepackage{float}
\usepackage{makecell}
\usepackage[most]{tcolorbox}
\usepackage{enumitem}
\usepackage{float}
\newcommand{\la}{\color{gray!120}}

\title{DIPrune: Task-Aware Token Pruning with Dual Importance for Efficient Multimodal Language Models}

\author{%
\textbf{Shuo Yang}$^{1*}$\quad
\textbf{Changbai Li}$^{1*}$\quad
\textbf{Linlin Yang}$^{2}$\quad
\textbf{Huobin Tan}$^{1}$\quad
\textbf{Rongyu Chen}$^{3}$\\
\textbf{Tongfei Chen}$^{1}$\quad
\textbf{Tian Wang}$^{1}$\quad
\textbf{Sheng Xu}$^{2}$\quad
\textbf{Baochang Zhang}$^{1}$\\[3pt]
$^{1}$Beihang University\quad
$^{2}$Communication University of China\\
$^{3}$National University of Singapore
}
\iclrfinalcopy

\begin{document}

\maketitle
\lhead{Preprint.}

{\renewcommand{\thefootnote}{\fnsymbol{footnote}}
\footnotetext[1]{Equal contribution.}}

\begin{abstract}
Recent training-free pruning approaches for Multimodal Large Language Models (MLLMs) effectively cut computational overhead by exploiting visual redundancy or text-vision attention. However, they frequently suffer from  semantic degradation due to their task-agnostic design or unreliable attention estimates. 
Based on our empirical analysis, we have found that this issue arises because salient tokens in shallow layers persistently suppress emerging semantic ones through numerical inertia, leading to premature discarding of signals crucial for deep reasoning. 
To address the 
aforementioned issue, from the task-oriented aspects, we first reformulate training-free pruning as a minimization of the distortion in the final task loss and derive a tractable, token-wise upper bound to serve as a surrogate objective. Specifically, this formulation inherently reveals a previously neglected inter-layer term that accounts for gradients across layers.
Accordingly, for the implementation, we propose \textbf{DIPrune}, a rank-based framework that employs a dual importance scoring mechanism to jointly optimize intra-layer static feature saliency and inter-layer dynamic semantic evolution. 
Extensive experiments on LLaVA and Qwen-VL demonstrate that DIPrune consistently achieves state-of-the-art results.

\end{abstract}

\section{Introduction}
\label{sec:intro}

Recent advances in Multimodal Large Language Models (MLLMs) have achieved strong performance but face a critical computational bottleneck: high-resolution images~\citep{li2025mini} produce an excessive number of visual tokens, leading to quadratic growth in self-attention cost. Training-free token pruning~\citep{bolya2022token} has emerged to address this, with current methods falling into two categories. Visual-redundancy-based pruning~\citep{yang2025visionzip,wen2025stop,zou2026don,zhang2025vscan} removes statistically similar tokens but operates independently of the text instruction, lacking semantic guidance and risking the loss of {task-relevant content}. Text-vision-interaction-based pruning~\citep{chen2024image, zhang2024sparsevlm, xing2025conical} uses cross-modal attention for guidance, yet remains susceptible to inherent attention biases, such as position bias~\citep{endo2025feather,zhang2024cls}, which can favor spatially proximate over semantically important tokens. Both paradigms therefore exhibit distinct limitations, either being task-agnostic or relying on unreliable attention estimates. 
It is, therefore, non-trivial to achieve task-aware and reliable attention estimates for training-free token pruning.

\begin{figure*}[t]
  \centering
  \includegraphics[
    width=1.0\textwidth, 
    trim=1 1 1 1,
    clip
  ]{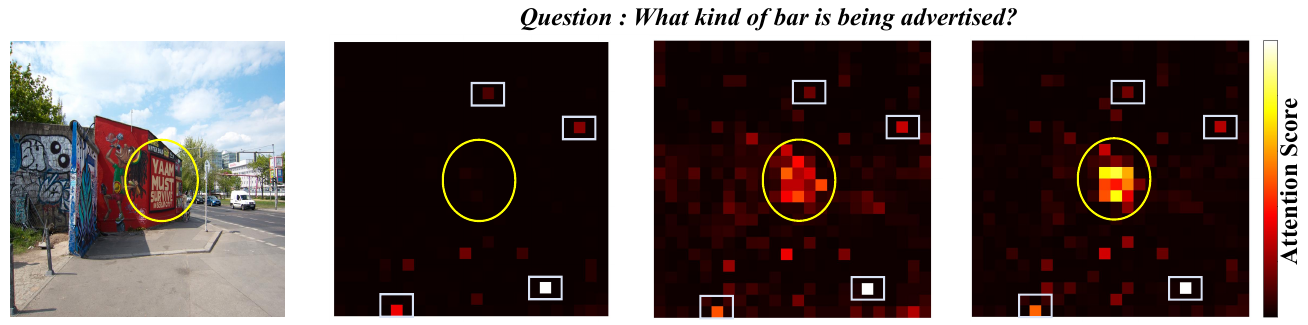}
    \caption{{Cross-Layer Attention Evolution.}
    Attention maps across depth (i.e., Layer 2/8/12) exhibit two behaviors: the answer-critical region (yellow circle) is weakly activated in early layers but gains attention with depth, while {boxed tokens (white rectangle)} remain highly attended across layers.}
  \label{fig:analyse3}
\end{figure*}

{
Empirically, we begin by conducting analysis and observation of visual tokens from the perspectives of different layers and tasks. Interestingly, we highlight an overlooked phenomenon for training-free token pruning: \textit{the Early Inertial Bias in attention distributions}, i.e., 
{tokens that acquire high attention weights in shallow layers exhibit strong numerical inertia, allowing them to persistently dominate in subsequent layers.
} This bias is exacerbated by the fact that the initial layers tend to focus on broad, generic visual features~\citep{shi2026vision, kang2025see}. Crucially, among these tokens, we distinguish a class that maintains high ranks solely due to inertia yet lacks task-driven dynamic evolution. We define these as \textbf{Inertial Tokens}, representing computational redundancy that should be discarded. 
{Consequently, prevailing pruning strategies, whether based on redundancy~\citep{shang2025llava,wang2025folder,arif2025hired,yang2025visionzip,zou2026don} or snapshot attention~\citep{xing2025conical,chen2024image,zhang2024sparsevlm,ye2025atp}, could rely excessively on these early, static importance signals.}
{This bias consistently leads to the premature removal of counterpart \textbf{Emerging Tokens}, i.e., tokens that appear less prominent in the shallow layers but increasingly capture semantically crucial information needed for task‑specific reasoning in deeper layers.}
}{To address the aforementioned limitations, we first formulate training-free pruning as a budget-constrained visual token pruning optimization problem, which minimizes the distortion in the final task loss. 
Since the original objective is intractable, we further derive a tractable, token-wise upper bound that we use as a surrogate objective. 

Crucially, this objective decouples into intra-layer and inter-layer terms, which respectively characterize Static Semantic Alignment through feature-salient importance and Dynamic Semantic Evolution through evolution-aware importance.  Especially, it inherently reveals a previously neglected inter-layer term that accounts for gradients across layers.

Accordingly,  we propose \textbf{DIPrune}, a rank-based framework for training-free token pruning. 
Based on our formulation, DIPrune instantiates a dual importance scoring mechanism that directly tackles the three core challenges: (1) To overcome the lack of textual guidance, it computes a {Semantic Alignment Score} ($S_{\text{align}}$) via cross-modal attention and inverse-rank normalization, prioritizing tokens semantically aligned with the instruction; (2) To mitigate reliance on biased attention estimates, it fuses $S_{\text{align}}$ and the {Semantic Evolution Score} ($S_{\text{evol}}$) with weights, providing a robust, multi-perspective importance assessment that avoids over-reliance on any single, potentially biased signal; (3) To counteract the early inertial bias, it calculates $S_{\text{evol}}$ by tracking rank displacement of tokens across adjacent layers, thereby preserving those gaining importance in deeper reasoning. 
By retaining only the top-$K$ tokens based on this dual importance score, DIPrune reduces the self-attention complexity while maintaining high semantic fidelity under aggressive compression.}

{Extensive experiments show that DIPrune demonstrates exceptional robustness and generalizability. Evaluations across diverse MLLM architectures further verify that our method consistently establishes new state‑of‑the‑art results.
The contributions of our work can be summarized as follows:
\begin{itemize}
    \item \textbf{Early Inertial Bias.} We systematically identify and validate the {Early Inertial Bias} in attention distributions, revealing that {shallow Inertial Tokens persistently suppress deep Emerging Tokens}.
    
    \item \textbf{DIPrune Framework.} We propose DIPrune, a rank-based pruning framework that employs a dual importance scoring mechanism to jointly optimize static semantic alignment and dynamic semantic evolution.
    
    \item \textbf{SOTA Performance.} DIPrune achieves new state-of-the-art results under high compression ratios across multiple mainstream MLLMs, demonstrating strong generalization capability and inference efficiency.
\end{itemize}
}

\section{Related Work}

\noindent\textbf{Multimodal Large Language Models.}
In recent years, Multimodal Large Language Models (MLLMs) have achieved remarkable progress~\citep{chen2024internvl, lin2024video, li2024llava, tong2024cambrian,zhu2024minigpt}. 
Evolving from early paradigms based on frozen components, such as BLIP~\citep{li2022blip} and Flamingo~\citep{alayrac2022flamingo}, to end-to-end training architectures like LLaVA~\citep{liu2023visual} and Qwen-VL~\citep{bai2025qwen25vltechnicalreport}, MLLMs have demonstrated superior performance across a variety of vision-language tasks. 
However, these performance gains are accompanied by substantial computational costs. High-resolution visual inputs induce an explosive growth in the number of visual tokens; for instance, while LLaVA-v1.5~\citep{liu2023visual} processes only 576 visual tokens, this number surges to 2,880 in LLaVA-NeXT~\citep{li2024llava}, with video models like Video-LLaVA~\citep{lin2024video} facing even more massive scales. 
Given that the computational complexity of the attention mechanism grows quadratically with sequence length, these massive visual sequences constitute a primary bottleneck for inference efficiency. 

\noindent\textbf{Vision Token Pruning.} To alleviate computational bottlenecks, recent research has explored various paradigms for visual token compression and pruning, which can be broadly categorized into two streams. The first category evaluates token importance using attention maps within Transformer layers. FastV~\citep{chen2024image} pioneers the use of attention mechanisms to identify and discard ineffective visual tokens. 
Building on this, PyramidDrop~\citep{xing2025conical} introduces a hierarchical pyramid structure, enabling progressive sequence compression. Furthermore, SparseVLM~\citep{zhang2024sparsevlm} leverages key text queries to guide the pruning process.
The second category focuses on eliminating inherent redundancy within visual features. VisionZip~\citep{yang2025visionzip} merges spatially redundant tokens based on visual feature similarity.
DART~\citep{wen2025stop} proposes a de-duplication mechanism based on representative pivots, retaining tokens with unique information. 
HoloV~\citep{zou2026don} introduces a spatial-aware budget allocation strategy to address background loss. Constrained by attention bias or the lack of textual semantic guidance, existing methods struggle to reconcile critical information retention with complex task adaptation. Instead, we propose a strategy that enhances efficiency while maintaining high semantic fidelity.

\section{Motivation and Analysis}
\label{sec:motivation}
In this section, we conduct an empirical analysis to investigate the impact of token importance across layers based on current attention-based pruning paradigms, and reveal the phenomenon of Early Inertial Bias inherent in attention distributions.

\begin{wrapfigure}{r}{0.5\linewidth}
    \centering
    \vspace{-0.3cm}
    \includegraphics[width=\linewidth]{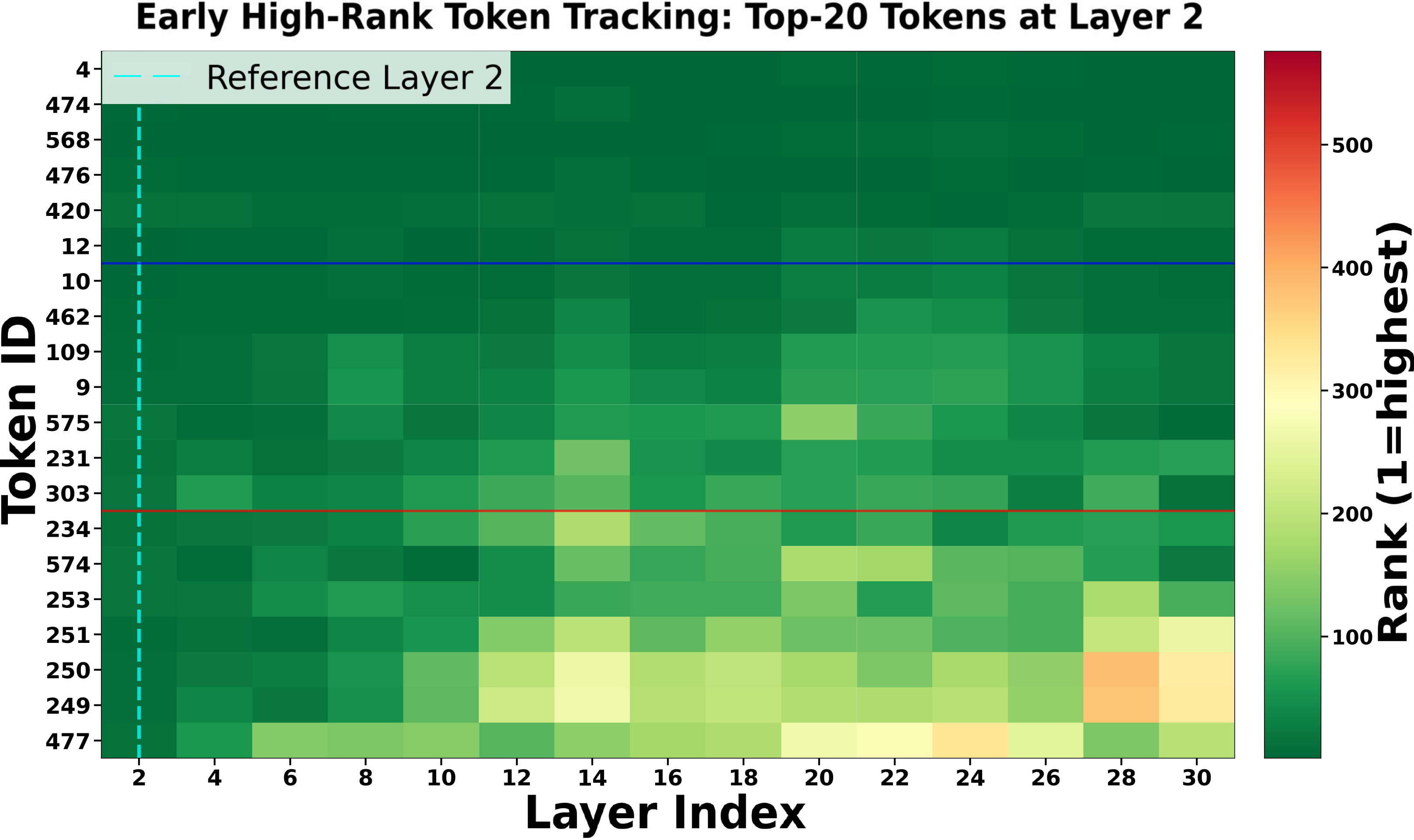}
    \caption{Analysis of Numerical Inertia. Visualization of the attention score and rank evolution of the Top-20 tokens from the second layer, please refer to Fig.~\ref{fig:analyse_8} for more details.}
    \label{fig:analyse1}
    \vspace{-0.3cm}
\end{wrapfigure}

\noindent\textbf{Observation I: {Inertial Bias Prioritizes Irrelevant Inertial Tokens}.} 
To verify the existence of Early Inertial Bias, we tracked the {averaged text-to-visual attention scores} and rank evolution of the Top-20 tokens from the second layer  to subsequent layers. 
As visualized in Figure~\ref{fig:analyse1}, {we can see that nearly half of the high-ranking tokens from the second layer maintain high ranks, while the rest gradually decrease in rank as the layer deepens}. 
Interestingly, for those tokens whose rank evolves progressively across layers, their importance are influenced by previous layers, exhibiting a form of inertia that prevents abrupt changes. For example, the Top-$K$ token sets significantly overlap between the second and the 10th layer, but not between the second and the 20th layer. Due to this {inertial bias}, existing attention-based pruning strategies tend to retain visually prominent yet potentially task-irrelevant generic features, leading to an inefficient allocation of the token budget toward {Inertial Tokens}.

\begin{wrapfigure}{r}{0.5\linewidth}
    \centering
    \vspace{-0.3cm}
    \includegraphics[
        width=\linewidth,
        trim=1 1 1 1,
        clip
    ]{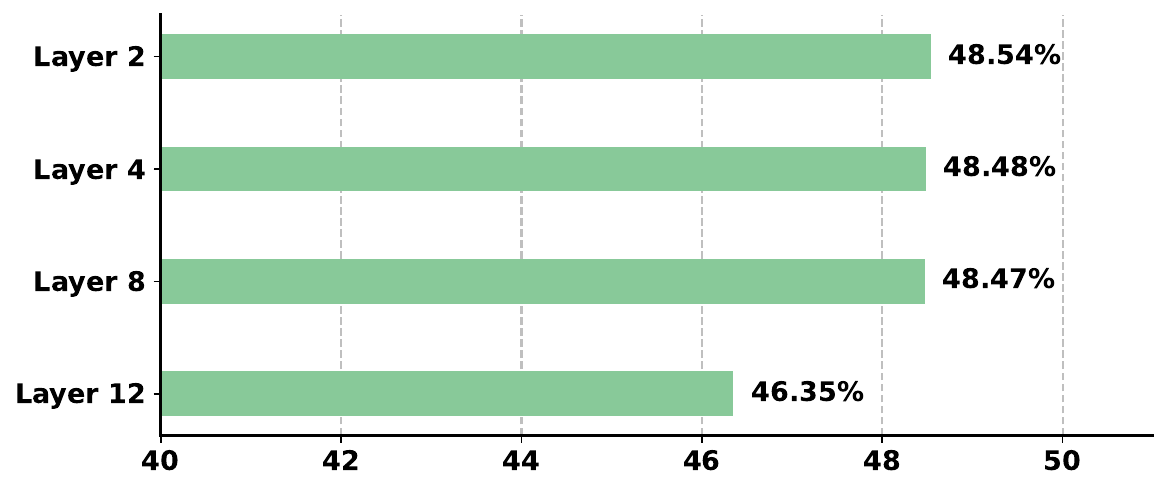}
    \caption{Analysis of Rank Discrepancy. Comparison between the final-layer key tokens and their rank distribution in early layers (2,4,8,12).}
    \label{fig:analyse2}
    \vspace{-0.3cm}
\end{wrapfigure}

\noindent\textbf{Observation II: Inertial Bias Masks Critical Emerging Tokens.} 
To assess the reliability of low attention scores in early layers, we compared the key token set from the final layer (32nd layer) against their rank distribution in early layers. As visualized in Figure~\ref{fig:analyse2}, a rank-value discrepancy is reflected by the statistic that approximately 47\% of the final-layer key tokens exhibit low scores in shallow layers.
This discrepancy confirms that, due to numerical inertia, many tokens critical for final reasoning are masked in early layers, resulting in low scores. 
This suggests that the limitation of the current static pruning perspective may lead to the information loss of these tokens with evolutionary potential.

This is also confirmed by the qualitative visualization of cross-layer attention (Figure~\ref{fig:analyse3}). Tracking the evolution from the second to the 12th layer clearly reveals the coexistence of two distinct behavioral patterns: regions associated with generic backgrounds often maintain high attention scores due to inertia, whereas critical regions pointing to the answer are not significantly activated in early layers but gradually accumulate importance as network depth increases.

\noindent\textbf{Remark.} The discrepancy in layer-wise behaviors reveals the limitation of existing strategies: an evaluation based solely on a single-layer static perspective struggles to differentiate between {Inertial Tokens and Emerging Tokens}. 
This suggests that the core of the evaluation of token importance should not rely solely on static observations of a specific layer but incorporate dynamic evolutionary trends relative to the final objective of the task.
This insight prompts us to extend our evaluation perspective to both intra-layer statistics and the inter-layer evolution. Consequently, in Section~\ref{sec:theory}, we will theoretically derive a pruning metric from a task-aware perspective, aiming to identify an effective indicator capable of rectifying inertial bias and capturing dynamic semantics.

\section{Theoretical Analysis}
\label{sec:theory}

\subsection{Optimization Objective under Budget Constraint}

In a standard MLLM, the visual encoder and text tokenizer project inputs into a unified sequence $\mathbf{X} \in \mathbb{R}^{(N+M) \times D}$, comprising $N$ visual tokens $\mathbf{V}$ and $M$ text tokens $\mathbf{T}$. Since the self-attention mechanism entails $\mathcal{O}((N+M)^2)$ complexity dominated by massive visual tokens, pruning visual tokens becomes essential for inference efficiency.

Focusing on the $l$-th Transformer layer, let $\mathbf{X}^{(l)} = [\mathbf{V}^{(l)}; \mathbf{T}^{(l)}]$ denote the multimodal input.

Since our pruning target is restricted to the visual modality, we focus on optimizing the binary mask $\mathbf{m} \in \{0, 1\}^N$ applied to $\mathbf{V}^{(l)}$.
Our goal is to find a mask $\mathbf{m}$ with a strict sparsity constraint $\|\mathbf{m}\|_0 = K$ that minimizes the distortion in the final task loss $\mathcal{L}$. Let $\mathcal{F}(\cdot)$ denote the mapping from layer $l$ to the final loss. The optimization objective is formulated as:
\begin{gather}
    \min_{\mathbf{m}} \Delta \mathcal{L} = \left| \mathcal{F}([\mathbf{m} \odot \mathbf{V}^{(l)}; \mathbf{T}^{(l)}]) - \mathcal{F}([\mathbf{V}^{(l)}; \mathbf{T}^{(l)}]) \right| \notag \\
    \text{s.t.} \quad \|\mathbf{m}\|_0 = K.
    \label{eq:objective}
\end{gather}
\textit{Remark: While formulated as a masking operation for theoretical derivation, in practice, unselected tokens are physically removed to accelerate inference.}

\subsection{Theoretical Error Decomposition}
\label{Theoretical Error Decomposition}
Direct optimization of Eq.~\ref{eq:objective} faces two primary challenges: 1) the discrete constraints imposed by the binary mask result in a complex discrete optimization problem; and 2) the exact gradient $\nabla \mathcal{L}$ is inaccessible during the forward pass. Consequently, we derive a tractable, token-wise upper bound for Eq.~\ref{eq:objective} to serve as a surrogate objective.

\paragraph{Derivation.}
Let the pruning perturbation for the $i$-th token be $\Delta \mathbf{v}_i = -(1-m_i)\mathbf{v}_i$. With the assumption that $\mathcal{F}$ is locally $L_{\mathcal{F}}$-smooth~\citep{nesterov2013introductory}, we apply the Descent Lemma for Eq.~\ref{eq:objective} and get the upper bound:

\begin{equation}
    \Delta \mathcal{L} \le \left| \langle \nabla_{\mathbf{V}^{(l)}} \mathcal{F}, \Delta \mathbf{V} \rangle \right| + \frac{L_{\mathcal{F}}}{2} \|\Delta \mathbf{V}\|^2_F,
\end{equation}
where $\Delta \mathbf{V} = (\mathbf{m} - \mathbf{1}) \odot \mathbf{V}^{(l)}$ denotes the aggregate perturbation matrix, $\nabla_{\mathbf{V}^{(l)}}\mathcal{F}$ denotes the gradient matrix, and $L_{\mathcal{F}}$ bounds the Hessian spectral norm. To derive a selection criterion for individual tokens, we further exploit the linearity of the inner product and the additivity of the Frobenius norm to decompose the upper bound at the level of individual tokens. Let $\mathbf{g}_i = \nabla_{\mathbf{v}_i} \mathcal{F}$ denote the gradient w.r.t. token $\mathbf{v}_i$. The error bound can be written as:
\begin{equation}
    \Delta \mathcal{L} \le \sum_{i=1}^N (1-m_i) \left( \left| \langle \mathbf{g}_i, \mathbf{v}_i \rangle \right| + \frac{L_{\mathcal{F}}}{2} \|\mathbf{v}_i\|^2 \right).
\end{equation}

\begin{figure*}[t]
    \centering
    \includegraphics[width=1.0\textwidth]{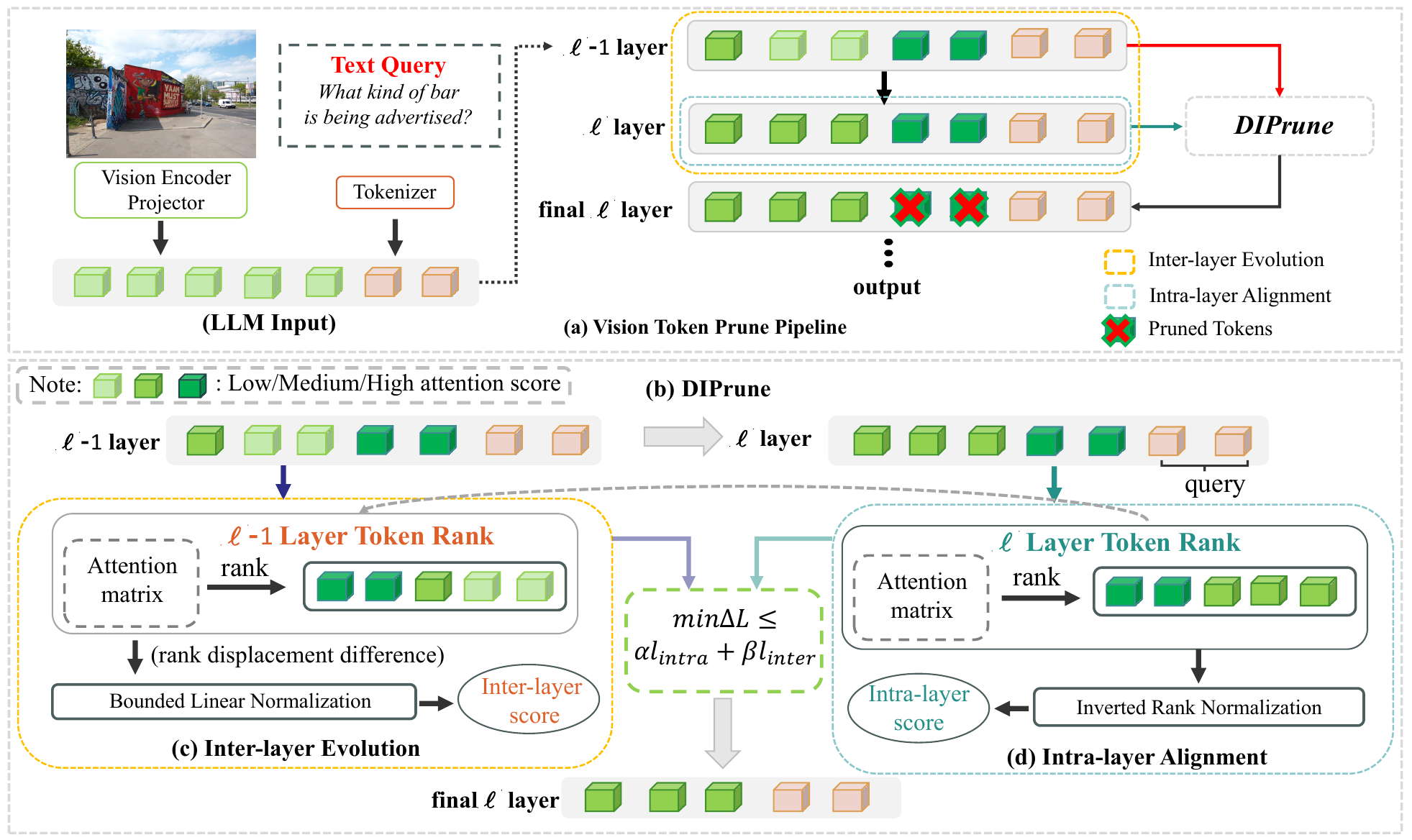}

        \caption{{Overview of the DIPrune Framework.} (a) The pipeline illustrates the multimodal encoding process (left) and the pruning operation at specific layers (right). (b-d) The DIPrune Mechanism: A dual importance scoring design that captures (c) Inter-layer Dynamic Semantic Evolution and evaluates (d) Intra-layer Static Feature Saliency. The two scores are jointly used to select the optimal top-$K$ tokens.}    
    
    \label{fig:framework} 
    \vspace{-10pt}
\end{figure*}

Based on the Cauchy-Schwarz inequality, we bound the interaction term by the product of the gradient and perturbation norms, thus removing any dependence on their relative orientation.
Moreover, we employ Young’s Inequality with a positive coefficient $\mu$ to decouple these coupled factors into independent additive constraints. The bound is relaxed to:
\begin{equation}
    \Delta \mathcal{L} \le \sum_{i=1}^N (1-m_i) \left( \underbrace{ \frac{1}{2\mu} \|\mathbf{g}_i\|^2 }_{\mathcal{I}_{\text{Inter}}} + \underbrace{ \frac{\mu + L_{\mathcal{F}}}{2} \|\mathbf{v}_i\|^2 }_{\mathcal{I}_{\text{Intra}}} \right).
    \label{eq:token_bound}
\end{equation}
To render the tractable global gradient energy $\|\mathbf{g}_i\|^2$ in $\mathcal{I}_\text{Inter}$ computable during inference, we utilize a local truncated horizon $\delta$ to approximate the inter-layer dependencies. By the chain rule, the gradient flow can be approximated as $\mathbf{g}_i^{(l)} \approx (\frac{\partial \mathbf{V}^{(l+\delta)}}{\partial \mathbf{v}_i^{(l)}})^\top \mathbf{g}^{(l+\delta)}$. Here, the magnitude of the Jacobian $\|\frac{\partial \mathbf{V}^{(l+\delta)}}{\partial \mathbf{v}_i^{(l)}}\|$ quantifies the token's influence on the evolution of intermediate representations.

Eq.~\ref{eq:token_bound} effectively transforms the combinatorial global minimization problem into a tractable local selection task. Under the sparsity budget $K$, minimizing the error upper bound is achieved by retaining the $K$ tokens with the largest combined importance scores $w \triangleq \mathcal{I}_{\text{Inter}} + \mathcal{I}_{\text{Intra}}$. Notably, this decomposition inherently reveals two distinct importances, serving as the theoretical foundation for our dual importance framework:
\begin{itemize}
    \item \textbf{Inter-layer Importance ($\mathcal{I}_{\text{Inter}}$):} This term is governed by the per-token gradient energy $\|\mathbf{g}_i\|^2$. Deriving from the sensitivity of the global loss to the current token through subsequent layers, it inherently captures the inter-layer dependencies.
    
    \item \textbf{Intra-layer Importance ($\mathcal{I}_{\text{Intra}}$):} This term is determined by the per-token feature magnitude $\|\mathbf{v}_i\|^2$. As it depends solely on the information content of the removed features within the current layer, it represents the intra-layer information. 
\end{itemize}

This theoretical decomposition forms the basis of the design principles underlying our framework, which we describe in detail in the next section.

\section{DIPrune Framework}
\label{sec:method_dyprune}

\begin{table*}[t]
\centering

\caption{Performance Comparison on LLaVA-v1.5-7B across Diverse Benchmarks. Results are reported under different pruning ratios. Gray shading denotes our method, and \textbf{bold} indicates the best results. Gray-text rows ($\dagger$) apply baselines at the same pruning layer as DIPrune (L6/L10); DIPrune(Ours) prunes at L10 by default.}
\label{tab:performance_comparison}
\resizebox{\linewidth}{!}{%
\begin{tabular}{@{}lccccccccc@{}}
\toprule
\textbf{Methods} & \textbf{GQA} & \textbf{MMB} & \textbf{MMB\textsubscript{CN}} & \textbf{MME} & \textbf{POPE} & \textbf{SQA} & \textbf{VQA\textsubscript{v2}} & \textbf{VQA\textsubscript{Text}} & \textbf{Average} \\
\midrule\midrule
Upper Bound, 576 Tokens & 61.9 & 64.7 & 58.1 & 1862 & 85.9 & 69.5 & 78.4 & 58.2 & 100\% \\
\midrule
\multicolumn{10}{@{}l}{\textbf{LLaVA-1.5 7B \hfill Retain 192 Tokens ($\downarrow 66.7\%$)}} \\
\cmidrule(r){1-10}
FastV (ECCV'24) & 52.7 & 61.2 & 57.0 & 1612 & 64.8 & 67.3 & 67.1 & 52.5 & 89.4\% \\
PDrop (CVPR'25) & 57.1 & 63.2 & 56.8 & 1766 & 82.3 & 68.8 & 75.1 & 56.1 & 96.4\% \\
VisionZip (CVPR'25) & 59.3 & 64.5 & 57.3 & 1767 & 86.4 & 68.9 & 76.8 & 57.3 & 98.6\% \\
SparseVLM (ICML'25) & 57.6 & 62.5 & 53.7 & 1721 & 83.6 & 69.1 & 75.6 & 56.1 & 96.0\% \\
DART (EMNLP'25) & 58.9 & 63.6 & 57.0 & 1856 & 82.8 & \textbf{69.8} & 76.7 & 57.4 & 97.8\% \\
HoloV (NeurIPS'25) & 58.7 & \textbf{65.4} & 58.0 & 1771 & 85.0 & 67.7 & 76.4 & 55.9 & 97.6\% \\
\rowcolor{gray!15}DIPrune(Ours) & \textbf{61.2} & 64.2 & \textbf{58.7} & \textbf{1856} & \textbf{85.8} & 69.6 & \textbf{78.0} & \textbf{58.5} & \textbf{99.9\%} \\
\midrule
\multicolumn{10}{@{}l}{\textbf{LLaVA-1.5 7B \hfill Retain 128 Tokens ($\downarrow 77.8\%$)}} \\
\cmidrule(r){1-10}
FastV (ECCV'24) & 49.6 & 56.1 & 56.4 & 1490 & 59.6 & 60.2 & 61.8 & 51.3 & 83.8\% \\
PDrop (CVPR'25) & 56.0 & 61.1 & 56.6 & 1644 & 82.3 & 68.3 & 72.9 & 55.1 & 94.9\% \\
VisionZip (CVPR'25) & 57.6 & 63.4 & 56.7 & 1768 & 84.7 & 68.8 & 75.6 & 56.8 & 97.2\% \\
SparseVLM (ICML'25) & 56.0 & 60.0 & 51.1 & 1696 & 80.5 & 67.1 & 73.8 & 54.9 & 92.8\% \\
DART (EMNLP'25) & 57.9 & 63.2 & 57.0 & 1845 & 80.1 & 69.1 & 75.9 & 56.4 & 96.5\% \\
HoloV (NeurIPS'25) & 57.6 & 63.9 & 56.9 & 1771 & 82.2 & 69.3 & 75.4 & 55.6 & 96.5\% \\
\rowcolor{gray!15}DIPrune(Ours) & \textbf{60.5} & \textbf{64.1} & \textbf{58.9} & \textbf{1857} & \textbf{86.2} & \textbf{69.4} & \textbf{77.2} & \textbf{58.0} & \textbf{99.5\%} \\
\midrule
\multicolumn{10}{@{}l}{\textbf{LLaVA-1.5 7B \hfill Retain 64 Tokens ($\downarrow 88.9\%$)}} \\
\cmidrule(r){1-10}
FastV (ECCV'24) & 46.1 & 48.0 & 52.7 & 1256 & 48.0 & 51.1 & 55.0 & 47.8 & 74.4\% \\
PDrop (CVPR'25) & 41.9 & 33.3 & 50.5 & 1092 & 55.9 & 68.6 & 69.2 & 45.9 & 76.7\% \\
VisionZip (CVPR'25) & 55.1 & 60.1 & 55.4 & 1690 & 77.0 & 69.0 & 72.4 & 55.5 & 93.4\% \\
SparseVLM (ICML'25) & 52.7 & 56.2 & 46.1 & 1505 & 75.1 & 62.2 & 68.2 & 51.8 & 86.3\% \\
DART (EMNLP'25) & 55.9 & 60.6 & 53.2 & 1765 & 73.9 & \textbf{69.8} & 72.4 & 54.4 & 92.6\% \\
HoloV (NeurIPS'25) & 55.1 & \textbf{63.3} & 55.1 & 1703 & 76.9 & 69.2 & 72.7 & 54.9 & 93.7\% \\
\arrayrulecolor{gray!50}\cmidrule(r){1-10}\arrayrulecolor{black}
\la VisionZip$^{\dagger}_{\text{L6}}$  & \la 56.0 & \la 61.2 & \la 55.2 & \la 1729 & \la 78.2 & \la 69.1 & \la 73.2 & \la 55.0 & \la 93.9\% \\
\la HoloV$^{\dagger}_{\text{L6}}$      & \la 56.1 & \la 61.9 & \la 56.2 & \la 1717 & \la 75.6 & \la 69.1 & \la 73.1 & \la 56.1 & \la 94.0\% \\
\la DIPrune$_{\text{L6}}$              & \la 56.8 & \la 61.8 & \la 55.3 & \la 1744 & \la 81.1 & \la 67.7 & \la 73.3 & \la 55.3 & \la 94.6\% \\
\la VisionZip$^{\dagger}_{\text{L10}}$ & \la 57.3 & \la 62.3 & \la 55.8 & \la 1759 & \la 81.8 & \la 68.3 & \la 73.9 & \la 55.8 & \la 95.4\% \\
\la HoloV$^{\dagger}_{\text{L10}}$     & \la 57.5 & \la 62.5 & \la 56.4 & \la 1729 & \la 77.4 & \la 69.5 & \la 73.9 & \la 56.6 & \la 95.2\% \\
\arrayrulecolor{gray!50}\cmidrule(r){1-10}\arrayrulecolor{black}
\rowcolor{gray!15}DIPrune(Ours) & \textbf{58.9} & 62.9 & \textbf{56.7} & \textbf{1807} & \textbf{86.2} & 69.4 & \textbf{75.0} & \textbf{57.2} & \textbf{97.7\%} \\
\bottomrule
\end{tabular}%
}
\vspace{0.3cm}  

\caption{Performance Comparison on LLaVA-NeXT-7B across Diverse Benchmarks, retaining only 320 tokens. Gray shading denotes our method, and \textbf{bold} indicates the best results.}
\label{tab:performance_comparison_320}
\renewcommand{\arraystretch}{1.15}
\resizebox{\linewidth}{!}{%
\begin{tabular}{@{}lccccccccc@{}}
\toprule
\textbf{Methods} &
\textbf{GQA} & \textbf{MMB} & \textbf{MMB$_{\mathrm{CN}}$} &
\textbf{MME} & \textbf{POPE} & \textbf{SQA} &
\textbf{VQA$_{\mathrm{v2}}$} & \textbf{VQA$_{\mathrm{Text}}$} & \textbf{Average} \\
\midrule
Upper Bound, 2880 Tokens & 64.2 & 67.4 & 60.6 & 1851 & 86.5 & 70.1 & 81.8 & 64.9 & 100\% \\
\midrule
\multicolumn{10}{@{}l@{}}{\textbf{LLaVA-NeXT 7B \hfill Retain 320 Tokens} ($\downarrow$ 88.9\%)} \\
\midrule
FastV (ECCV'24) & 55.9 & 61.6 & 51.9 & 1661 & 71.7 & 62.8 & 71.9 & 55.7 & 87.2\% \\
PDrop (CVPR'25) & 56.4 & 63.4 & 56.2 & 1663 & 77.6 & 67.5 & 73.5 & 54.4 & 90.6\% \\
MustDrop (2024.11) & 57.3 & 62.8 & 55.1 & 1641 & 82.1 & 68.0 & 73.7 & \textbf{59.9} & 92.5\% \\
FasterVLM (arXiv'24) & 56.9 & 61.6 & 53.5 & 1701 & 83.6 & 66.5 & 74.0 & 56.5 & 91.1\% \\
SparseVLM (ICML'25) & 56.1 & 60.6 & 54.5 & 1533 & 82.4 & 66.1 & 71.5 & 58.4 & 90.6\% \\
DART (EMNLP'25) & 61.7 & 65.3 & 58.2 & 1710 & 84.1 & 68.4 & 79.1 & 58.7 & 95.9\% \\
HoloV (NeurIPS'25) & 61.7 & 65.3 & 57.5 & 1738 & 83.9 & 68.9 & 79.5 & 58.7 & 95.8\% \\
\rowcolor{gray!15}DIPrune (ours) & \textbf{63.1} & \textbf{67.9} & \textbf{60.4} & \textbf{1788} & \textbf{87.7} & \textbf{69.3} & \textbf{80.3} & 58.7 & \textbf{98.2\%} \\
\bottomrule
\end{tabular}%
}
\vspace{-0.7cm}   
\end{table*}

Inspired by Eq.~\ref{eq:token_bound}, which describes the pruning error bound  in terms of two distinct importance, we introduce DIPrune, a rank-based framework as illustrated in Figure~\ref{fig:framework}.
To minimize the error bound, we formulate the token importance scoring function of $l$-th layer $S^{(l)}_\text{DIPrune}$ as a weighted fusion:
\begin{equation}
    S^{(l)}_\text{DIPrune}(\mathbf{v}_i) = \underbrace{\alpha \cdot S^{(l)}_\text{align}(\mathbf{v}_i)}_{ \mathcal{I}_\text{Intra}} + \underbrace{\beta \cdot S^{(l)}_\text{evol}(\mathbf{v}_i)}_{\mathcal{I}_\text{Inter}},
    \label{eq:total_score}
\end{equation}
where $S^{(l)}_\text{align}$ and $S^{(l)}_\text{evol}$ are the scores for {feature-salient importance (i.e., Semantic Alignment Score) and dynamic evolution importance (i.e., Semantic Evolution Score)}, respectively, and $\alpha, \beta$ are hyper-parameters with $\alpha+\beta=1$, balancing the trade-off between minimizing static information loss and preserving dynamic evolution potential.
By retaining the Top-$K$ tokens with the highest $S^{(l)}_\text{DIPrune}$, we effectively minimize the theoretical error upper bound derived in Section~\ref{Theoretical Error Decomposition}. 

\subsection{Semantic Alignment Score}

To mitigate the intra-layer information loss ($\mathcal{I}_{\text{Intra}}$), we prioritize tokens with high feature saliency. We utilize the cross-modal attention as a direct proxy for this saliency. Specifically, we define the saliency of visual tokens based on the attention scores where text tokens act as Queries ($\mathbf{Q}_T$) and visual tokens as Keys ($\mathbf{K}_V$). Let $\mathbf{A} \in \mathbb{R}^{M \times N}$ be the cross-modal attention matrix averaged across all heads. The raw importance $a_i$ for each visual token $\mathbf{v}_i$ is computed by aggregating attention weights across all text queries:
\begin{equation}
    a_i = \frac{1}{M} \sum_{j=1}^{M} \mathbf{A}_{j,i}.
    \label{eq:raw_importance}
\end{equation}
To ensure scale invariance across layers, we employ Inverted Rank Normalization on $\{a_i\}_{i=1}^N$. Defining $\mathcal{R}^{(l)}(\mathbf{v}_i) \in \{1, \dots, N\}$ as the rank of token $\mathbf{v}_i$ based on $a_i$, the Semantic Alignment Score is formulated as:
\begin{equation}
    S_\text{align}^{(l)}(\mathbf{v}_i) = \frac{N - \mathcal{R}^{(l)}(\mathbf{v}_i) + 1}{N}.
    \label{eq:align_importance}
\end{equation}
This metric maps importance to the $(0, 1]$ interval, providing a robust measure of static alignment independent of absolute attention magnitudes.
\subsection{Semantic Evolution Score}

To capture the Dynamic Semantic Evolution ($\mathcal{I}_{\text{Inter}}$), we focus on tokens exhibiting significant semantic shifts across layers. Instead of relying on noisy feature distances, we quantify the inter-layer rank displacement to capture the trajectory of importance. Defining the rank displacement over a step $\delta$ as $\Delta \mathcal{R}^{(l)}_i \triangleq \mathcal{R}^{(l-\delta)}(\mathbf{v}_i) - \mathcal{R}^{(l)}(\mathbf{v}_i)$, we calculate the Semantic Evolution Score via a Bounded Linear Normalization:
\begin{equation}
    S_\text{evol}^{(l)}(\mathbf{v}_i) = \frac{1}{2} \left( \frac{\text{Clamp}(\Delta \mathcal{R}^{(l)}_i, -M_\text{max}, M_\text{max})}{M_\text{max}} + 1 \right).
    \label{eq:s_evol}
\end{equation}
By clamping the rank displacement within $[-M_\text{max}, M_\text{max}]$, this metric identifies tokens
with rapidly rising priority while suppressing spurious fluctuations; the unselected tokens and their KV
caches are then removed, reducing attention cost to $\mathcal{O}((K{+}M)^2)$ (see Appendix~\ref{supp:complexity_analysis}).

\section{Experiments}
\subsection{Setup}
\noindent\textbf{Models.}
We evaluate DIPrune across diverse and representative MLLMs to ensure its generalizability and scalability under diverse architectural settings. 
Specifically, we adopt LLaVA-v1.5 (7B/13B) as the standardized multimodal benchmark backbone. 
We further include the high-resolution variant LLaVA-NeXT to examine performance under extremely dense visual token conditions. To verify cross-architecture universality and temporal adaptability, we extend our evaluation to Qwen2.5-VL, Video-LLaVA, and the latest Qwen3-VL series models, respectively.
Due to space constraints, detailed descriptions of the evaluation benchmarks and specific implementation details are provided in the  Appendix~\ref{sec:supp_benchmarks} and \ref{sec:supp_implementation}.

\subsection{Main Results}
As reported in Tab.~\ref{tab:performance_comparison}, DIPrune achieves the highest average performance
across all budgets on LLaVA-v1.5-7B. It retains 99.9\% of the unpruned performance at 192 tokens
($\downarrow$66.7\%) and exceeds the best baseline by 2.3\% at 128 tokens ($\downarrow$77.8\%). Even at
64 tokens ($\downarrow$88.9\%), it retains 97.7\%, comparable to HoloV with 192 tokens (97.6\%). Under
layer-aligned settings (gray rows), DIPrune maintains consistent advantages at both L6 and L10, with the
margin widening at the deeper layer, consistent with the inter-layer evolution signal becoming more
informative with depth.

\subsection{Generalization and Robustness}

While the results on LLaVA-1.5-7B validate the effectiveness of DIPrune, we further conduct a systematic evaluation under more challenging settings to assess its generalization across model scales, visual sequence lengths, architectures, and modalities. Specifically, we examine high-resolution inputs on LLaVA-NeXT-7B and cross-architecture transfer on the Qwen-VL series below, while results on model scaling (LLaVA-1.5-13B), video understanding (Video-LLaVA-7B), visual grounding, and generative tasks are deferred to the Appendix~\ref{supp:model_Generalization}.

\begin{wraptable}[18]{r}{0.48\linewidth}
    \centering
    \vspace{-14pt}      
    \caption{Performance Comparison on Qwen2.5-VL-7B.}
    \label{tab:comparative_experiments}
    \vspace{-4pt}

    \resizebox{\linewidth}{!}{
        \setlength{\tabcolsep}{3.5pt}
        \renewcommand{\arraystretch}{1.1}
        \begin{tabular}{lccccc}
        \toprule
        \textbf{Methods} & \textbf{MMB} & \textbf{MME} & \textbf{POPE} & \textbf{SQA} & \textbf{Avg.} \\
        \midrule
        Upper Bound & 82.8 & 2304 & 86.1 & 84.7 & 100\% \\
        \midrule
        \multicolumn{6}{c}{\textit{Token Pruning Rate = 66.7\%}} \\
        \midrule
        FastV & 75.7 & 2072 & 82.2 & 78.5 & 92.6\% \\
        HoloV & 78.3 & 2093 & 85.0 & 79.8 & 94.6\% \\
        \rowcolor{gray!15} DIPrune (Ours) & \textbf{81.4} & \textbf{2193} & \textbf{87.5} & \textbf{80.8} & \textbf{97.6\%} \\
        \midrule
        \multicolumn{6}{c}{\textit{Token Pruning Rate = 77.8\%}} \\
        \midrule
        FastV & 74.9 & 2036 & 80.7 & 78.0 & 91.2\% \\
        HoloV & 76.5 & 2043 & 82.3 & 79.8 & 92.7\% \\
        \rowcolor{gray!15} DIPrune (Ours) & \textbf{80.2} & \textbf{2114} & \textbf{85.4} & \textbf{80.5} & \textbf{95.7\%} \\
        \midrule
        \multicolumn{6}{c}{\textit{Token Pruning Rate = 88.9\%}} \\
        \midrule
        FastV & 69.2 & 1940 & 78.6 & 77.4 & 87.6\% \\
        HoloV & 72.4 & 2006 & 80.7 & 79.5 & 90.5\% \\
        \rowcolor{gray!15} DIPrune (Ours) & \textbf{76.1} & \textbf{2053} & \textbf{84.2} & \textbf{79.9} & \textbf{93.3\%} \\
        \bottomrule
        \end{tabular}
    }
\end{wraptable}

\noindent\textbf{High-Resolution Adaptation: LLaVA-NeXT-7B.} To evaluate DIPrune under ultra-long visual token streams, we test it on LLaVA-NeXT-7B with an extreme compression setting ( $\downarrow$88.9\%). As shown in Table~\ref{tab:performance_comparison_320}, DIPrune achieves 98.2\% average performance retention, substantially outperforming FastV (87.2\%, +11.0\%) and SparseVLM (90.6\%, +7.6\%). These results demonstrate strong sequence-length robustness: DIPrune remains highly effective even with thousands of visual tokens under stringent budgets. 

\noindent\textbf{Cross-Architecture Generalization: Qwen-VL Series.}
To verify that DIPrune does not rely on LLaVA-specific designs, we apply the same pruning pipeline to
Qwen2.5-VL-7B, which differs substantially in its native dynamic-resolution visual encoding and multimodal
positional scheme. As shown in Table~\ref{tab:comparative_experiments}, DIPrune consistently outperforms FastV
and HoloV across all token budgets, surpassing HoloV by about 3\% in average performance at each pruning ratio
(e.g., 93.3\% vs.\ 90.5\% at 88.9\%), with notable gains on MMBench (+3.7) and POPE (+3.5). Moreover,
consistent improvements on Qwen3-VL Dense-8B and MoE-30B-A3B (Appendix~\ref{supp:model_Generalization}) further
confirm that DIPrune generalizes across both dense and MoE architectures.

\begin{figure}[t]
    \centering
    \begin{subfigure}{0.47\linewidth}
        \centering
        \includegraphics[width=\linewidth]{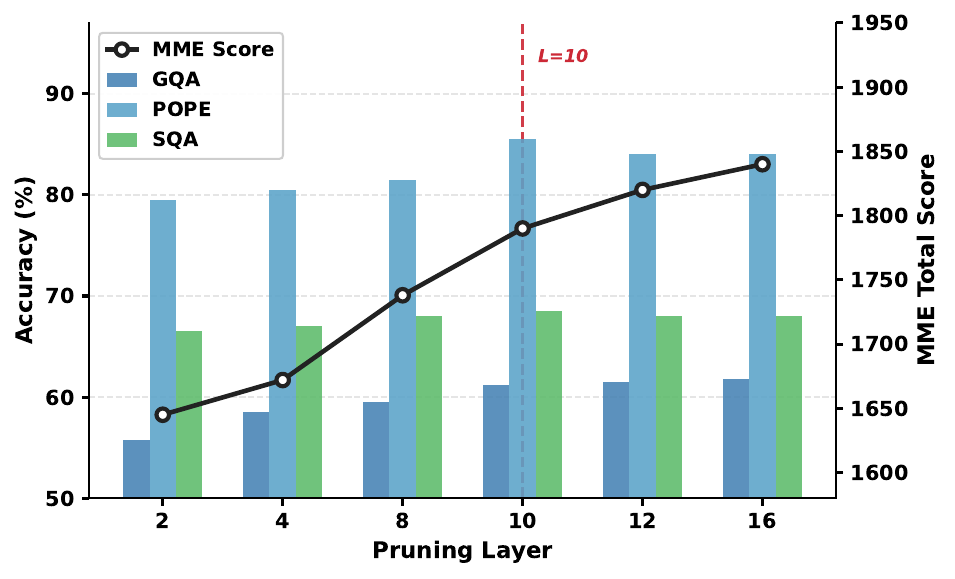}
        \caption{Impact of Pruning Layer $L$ on LLaVA-NeXT-7B.}
        \label{fig:ablation_layer_span}
    \end{subfigure}
    \hfill
    \begin{subfigure}{0.49\linewidth}
        \centering
        \includegraphics[width=\linewidth]{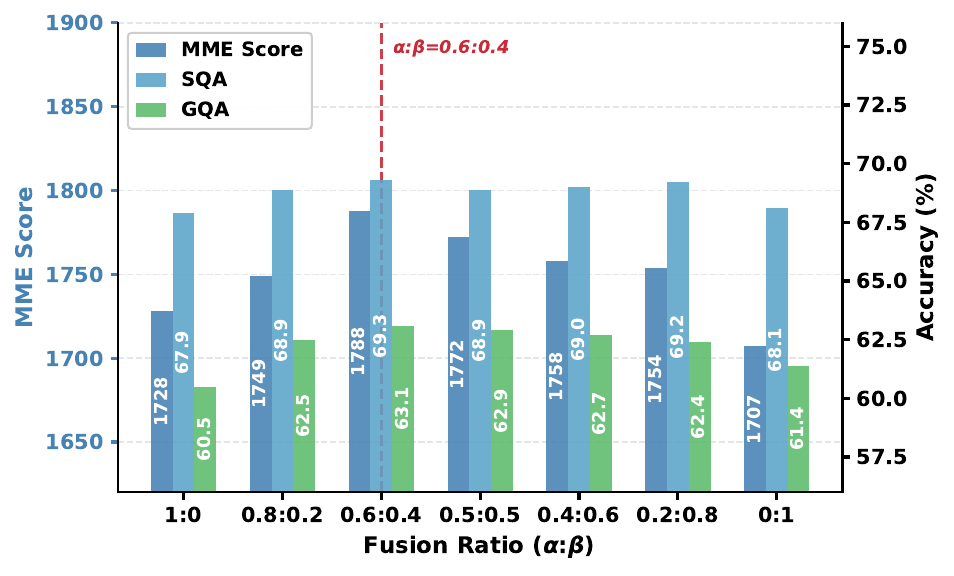}
        \caption{Sensitivity of Fusion Ratio($\alpha$:\,$\beta$).}
        \label{fig:ablation_alpha_beta}
    \end{subfigure}
    \vspace{-0.3cm}

    \caption{\textbf{Ablation studies} on LLaVA-NeXT-7B.}

    \vspace{-0.5cm}
\end{figure}

\subsection{Ablation Studies}
\textbf{Effectiveness of Pruning Layers.}
The pruning depth dictates the trade-off between semantic preservation and computational efficiency. We evaluate DIPrune by varying the pruning layer $L \in \{2, 4, 8, 10, 12, 16\}$. As illustrated in Figure~\ref{fig:ablation_layer_span}, performance consistently improves as the pruning stage moves deeper. Specifically, early-stage pruning (e.g., $L \le 4$) results in substantial degradation, as multimodal embeddings require sufficient layers for stable feature abstraction and cross-modal integration. While deeper layers (e.g., $L \ge 12$) offer marginal gains, they yield smaller computational savings due to the massive visual token count. Consequently, we select the $10$-th layer as the default pruning layer to strike an optimal balance between semantic integrity and inference throughput. Please refer to Appendix~\ref{supp:model_ablation} for more details.

\textbf{Effectiveness of Dual Importance.}
On LLaVA-NeXT-7B, we observe optimal performance with the configuration of $\alpha = 0.6$ and $\beta = 0.4$ (Figure~\ref{fig:ablation_alpha_beta}). This equilibrium precisely highlights the functional trade-off: an excessively high $\alpha$ ignores the inter-layer semantic dynamics, while an over-reliance on $\beta$ introduces instability due to the lack of saliency anchoring from the current layer. The consistent optimality of this ratio across diverse architectures confirms the robustness of our design.
Furthermore, we conduct additional sensitivity analysis experiments on Qwen2.5-VL-7B, which consistently validate the stability of the dual importance parameter configuration.

\subsection{Efficiency Analysis}
\label{sec:efficiency}

\begin{table}[t]
\centering
\vspace{4pt}
\caption{\textbf{Efficiency analysis} on LLaVA-1.5-7B (64 tokens, 88.9\% pruning). TTFT (ms) and ITT
(ms/token) are measured on POPE. (a) Default configurations; (b) baselines pruned at the same layer as
DIPrune. Acc: average relative performance.}
\label{tab:efficiency}
\vspace{-6pt}
\small
\setlength{\tabcolsep}{4pt}
\begin{minipage}[t]{0.485\linewidth}
\centering
\textit{(a) Default configurations}\\[2pt]
\begin{tabular*}{\linewidth}{@{\extracolsep{\fill}}lccc@{}}
\toprule
Method & TTFT$\downarrow$ & ITT$\downarrow$ & Acc$\uparrow$ \\
\midrule
Baseline   & 73.23 & 16.21 & 100\% \\
FastV      & 32.43 & 14.95 & 74.4\% \\
SparseVLM  & 33.43 & 14.98 & 86.3\% \\
HoloV      & 31.83 & 14.94 & 93.7\% \\
\rowcolor{gray!15} DIPrune & 32.50 & 14.96 & \textbf{97.7\%} \\
\midrule
\textit{Overhead} & \multicolumn{3}{c}{0.24\,ms (0.74\% of TTFT)} \\
\bottomrule
\end{tabular*}
\end{minipage}\hfill
\begin{minipage}[t]{0.485\linewidth}
\centering
\textit{(b) Layer-aligned comparison}\\[2pt]
\begin{tabular*}{\linewidth}{@{\extracolsep{\fill}}lccc@{}}
\toprule
Method & TTFT$\downarrow$ & Speedup$\uparrow$ & Acc$\uparrow$ \\
\midrule
VisionZip$\smash{^{\dagger}_{\text{L6}}}$ & 29.04 & 2.52$\times$ & 93.9\% \\
HoloV$\smash{^{\dagger}_{\text{L6}}}$     & 32.13 & 2.28$\times$ & 94.0\% \\
\rowcolor{gray!15} DIPrune$\smash{_{\text{L6}}}$ & 31.58 & 2.32$\times$ & \textbf{94.6\%} \\
\midrule
VisionZip$\smash{^{\dagger}_{\text{L10}}}$ & 30.57 & 2.40$\times$ & 95.4\% \\
HoloV$\smash{^{\dagger}_{\text{L10}}}$     & 33.83 & 2.16$\times$ & 95.2\% \\
\rowcolor{gray!15} DIPrune$\smash{_{\text{L10}}}$ & 32.50 & 2.25$\times$ & \textbf{97.7\%} \\
\bottomrule
\end{tabular*}
\end{minipage}
\vspace{-10pt}
\end{table}
As shown in Tab.~\ref{tab:efficiency}(a), DIPrune reduces TTFT by 55.6\% while ITT remains nearly
identical across pruning methods. At comparable TTFT, it retains notably higher accuracy (97.7\% vs.
93.7\% of HoloV) with only 0.24\,ms overhead (0.74\% of TTFT). To control for the pruning position,
Tab.~\ref{tab:efficiency}(b) evaluates VisionZip, HoloV, and DIPrune under identical pruning layers,
where DIPrune still leads by +0.6\% at L6 and +2.3\% at L10, indicating that the gain stems from the
token selection criterion (Appendix~\ref{app:eff}).
\section{Conclusion \& Limitations}
\label{sec:conclusion}
In this work, we identify the Early Inertial Bias in attention-based pruning for MLLMs and propose DIPrune, a rank-based framework with rigorous theoretical error bound decomposition. By evaluating both intra-layer static alignment and inter-layer dynamic evolution, DIPrune effectively preserves task-critical tokens. We note that in highly cluttered scenes or when text queries are extremely short, the margin of alignment-based pruning may shrink; furthermore, as its importance estimation is grounded in the semantic evolution within the LLM backbone, latency benefits may be constrained in scenarios that favor pre-LLM token compression.

\bibliography{refs}
\bibliographystyle{preprint}
\appendix
\newpage
\appendix
\onecolumn

\begin{tcolorbox}[
  enhanced, colback=blue!2!white, colframe=black!25, arc=3mm, boxrule=0.6pt,
  left=8pt, right=8pt, top=6pt, bottom=6pt
]
\begin{center}
{\Large\textbf{Supplementary Material}}
\end{center}
\vspace{-2pt}
\noindent\textbf{This appendix is organized as follows:}
\begin{itemize}[leftmargin=2.2em, itemsep=1pt, topsep=3pt]
  \item \textbf{Appendix~\ref{sec:Detailed}} details the experimental settings, including benchmarks,
        implementation details, computational complexity, the robustness of the query-window strategy,
        and the implementation of decoupled attention.
  \item \textbf{Appendix~\ref{supp:model_Generalization}} extends the evaluation to model scaling, video understanding,
        visual grounding, and the Qwen3-VL series, together with generative tasks and a stratified analysis
        on query length and scene density.
  \item \textbf{Appendix~\ref{supp:model_ablation}} presents further ablations, including the empirical validation
        of proxy scores, hyperparameter sensitivity, pruning-layer selection, and extended token budgets.
  \item \textbf{Appendix~\ref{app:eff}} provides a detailed efficiency analysis, including the measurement
        setup, overhead breakdown, layer-aligned comparison, and end-to-end deployment efficiency.
  \item \textbf{Appendix~\ref{app:visualization}} visualizes the retained visual tokens under different pruning ratios.
\end{itemize}
\end{tcolorbox}

\section{Detailed Experiments Settings}
\label{sec:Detailed}

\subsection{Benchmarks}
\label{sec:supp_benchmarks}
To comprehensively assess the impact of DIPrune on the core capabilities of MLLMs, we employ a diverse suite of eight widely recognized public benchmarks. 
Following standard evaluation protocols~\citep{shang2025llava, singh2019towards}, we categorize these benchmarks into four distinct competency domains: 
1) General Visual Question Answering, evaluated via VQAv2~\citep{goyal2017making} and GQA~\citep{hudson2019gqa}; 
2) Fine-grained Perception and Reasoning, assessed using TextVQA~\citep{singh2019towards} and ScienceQA~\citep{lu2022learn}; 
3) Hallucination Evaluation, gauged by POPE~\citep{li2023evaluating}; and 4) Comprehensive Instruction Following,
measured through MME~\citep{fu2026mme}, MMBench, and MMBench-CN~\citep{liu2024mmbench}.
Beyond these benchmarks, we further evaluate DIPrune on
(i) visual grounding with the RefCOCO series datasets~\citep{yu2016modelingcontextreferringexpressions} (RefCOCO, RefCOCO+, and RefCOCOg);
(ii) video question answering with MSVD-QA and MSRVTT-QA; and
(iii) generative and fine-grained tasks, including detailed captioning (DetailCaps~\citep{dong2024benchmarking}),
open-ended hallucination (CHAIR~\citep{rohrbach2018object}), multi-turn reasoning
(ConvBench~\citep{liu2024convbench}), OCR (OCRBench~\citep{liu2024ocrbench}), and chart understanding
(ChartQA~\citep{masry2022chartqa}).

\subsection{Implementation Details}
\label{sec:supp_implementation}
We implement DIPrune across a diverse set of MLLMs to validate its generalizability, including LLaVA-v1.5-7B/13B, LLaVA-NeXT-7B, Video-LLaVA-7B, and the Qwen-VL series (including Qwen2.5-VL-7B, Qwen3-VL-Dense-8B, and Qwen3-VL-MoE-30B), which span the two most representative multimodal architectures. Unless otherwise stated, all experiments strictly follow the official evaluation protocols and default scripts released with each model to ensure reproducibility and fair comparison. 

We default to performing one-shot pruning at the 10th layer of the LLM backbone. During the forward pass, we introduce a manual text-to-vision attention calculation (as formulated in Eq.~\ref{eq:raw_importance}). This design ensures that our pruning strategy is fully compatible with high-performance attention kernels such as SDPA and FlashAttention-2, thereby enabling more efficient inference computation without compromising the effectiveness of the strategy. For DIPrune, hyperparameters are set based on comprehensive ablation studies, with default values of $\delta = 1$, $M_{max} = 200$, $\alpha = 0.6$, and $\beta = 0.4$. This configuration yields consistent performance gains across various MLLMs and evaluation benchmarks, demonstrating robust cross-architecture applicability and stability.

\subsection{Computational Complexity Analysis}
\label{supp:complexity_analysis}
\noindent\textbf{Theoretical FLOPs Reduction.} 
Adopting the evaluation framework from PyramidDrop~\citep{xing2025conical}, we estimate the theoretical floating-point operations (FLOPs) incurred by the LLM decoding layers when processing visual features. The computational load across $K$ layers is primarily distributed between the causal self-attention blocks and the Feed-Forward Networks (FFN). The aggregate FLOPs for the pre-filling stage are calculated as follows:
\begin{equation}
    \text{FLOPs}_{\text{LLM}} = \sum_{k=1}^{K} \left( 4n_k d^2 + 2n_k^2 d + 3n_k d m \right),
    \label{eq:flops_total}
\end{equation}
where $n_k$ represents the number of visual tokens retained at layer $k$, while $d$ and $m$ denote the hidden state dimension and the FFN intermediate dimension, respectively. Crucially, the term $2n_k^2 d$ highlights the quadratic dependency of the attention mechanism on the token count. Therefore, strategically pruning $n_k$ provides a direct pathway to significantly alleviate the inference burden, offering substantial efficiency gains particularly for long-context multimodal inputs.

\vspace{0.2cm} 

\noindent\textbf{Computing Budget: Detailed Estimation of DIPrune.} 
To rigorously evaluate the efficiency of our proposed method, we provide a fine-grained estimation of the additional computational overhead introduced by DIPrune. The pruning process operates via three distinct stages:

\noindent\textbf{Lightweight Attention Construction.} 
In this stage, we generate a targeted importance map by computing the interaction between visual keys $\mathbf{K}_{vis}$ and a selected subset of text queries $\mathbf{Q}_{sub}$. To further minimize computational overhead, we implement a text query windowing strategy. Specifically, rather than performing a full-scale attention pass over the entire text sequence, we restrict the computation to a strategic window of the most recent text queries, typically set to $0.8\times$ the original text length (i.e., $L_{win} \approx 0.8 L_{total}$). Our empirical validation confirms that this truncated window yields performance virtually identical to the full-text baseline, while offering a tangible reduction in FLOPs (see Appendix~\ref{sec:window_ablation} 
for robustness analysis, and Appendix~\ref{supp:attn_detail} for implementation details of decoupled attention analysis).

This operation corresponds to a dense matrix multiplication between projections of shape $\mathbb{R}^{N_v \times D}$ and $\mathbb{R}^{D \times L_{win}}$. Consequently, the FLOPs for this tailored construction are reduced to:
\begin{equation}
    \text{FLOPs}_{\text{attn}} \approx 2 \cdot N_v \cdot L_{win} \cdot D
\end{equation}
where $N_v$ denotes the number of visual tokens and $D$ is the hidden state dimension. This formulation ensures that the overhead remains minimal and strictly proportional to the reduced query window size. Crucially, this decoupled design enables full compatibility with hardware-optimized attention kernels such as FlashAttention-2 and SDPA, which typically do not expose intermediate attention weights. By isolating the scoring computation, DIPrune retains the high throughput benefits of these efficient kernels for the primary forward pass.

\noindent\textbf{Dual Importance Scoring.} 
This stage transforms raw attention values into the final token importance score $S_{DIPrune}^{(l)}$. The computation consists of three sub-steps:
(1) \textit{Score Aggregation}: Averaging attention scores across $H$ heads and summing over $L_{win}$ queries involves approximately $N_v \cdot L_{win} \cdot H$ additions.
(2) \textit{Rank Calculation}: Deriving the rank $\mathcal{R}$ for $N_v$ tokens requires sorting, with a complexity governed by $N_v \log N_v$.
(3) \textit{Displacement \& Fusion}: Calculating the rank displacement ($\Delta \mathcal{R} = \mathcal{R}^{(l-\delta)} - \mathcal{R}^{(l)}$, consistent with Eq.~\ref{eq:s_evol}) and fusing it with the alignment score ($\alpha S_{align} + \beta S_{evol}$) involves element-wise operations on $N_v$ tokens, totaling approximately $5N_v$ FLOPs.
Therefore, the total FLOPs for the scoring stage can be approximated as:
\begin{equation}
    \text{FLOPs}_{\text{score}} \approx N_v \cdot (L_{win} \cdot H + \log N_v + 5)
\end{equation}
Given that $N_v$ (e.g., 576) is relatively small compared to the hidden dimension $D$ (e.g., 4096), this term is computationally negligible compared to the matrix multiplication in the previous stage.

\noindent\textbf{Pruning Execution.} 
The final stage involves selecting the top-$k$ indices and gathering the retained tokens. The selection process on an $N_v$-dimensional score vector has a linear complexity of $O(N_v)$. While the subsequent token gathering involves memory copy operations, the arithmetic FLOPs are nominally zero. Thus, the cost is bounded by:
\begin{equation}
    \text{FLOPs}_{\text{exec}} \approx N_v
\end{equation}

\noindent\textbf{Summary of Overhead.} 
Aggregating the components above, the total computational overhead introduced by DIPrune at the pruning layer is dominated by the lightweight attention construction:
\begin{align}
    \text{FLOPs}_{\text{total}} &= \text{FLOPs}_{\text{attn}} + \text{FLOPs}_{\text{score}} + \text{FLOPs}_{\text{exec}} \nonumber \\ 
    &\approx 2 \cdot N_v \cdot L_{win} \cdot D + O(N_v \cdot H) \nonumber \\
    &\approx 2 \cdot N_v \cdot L_{win} \cdot D
\end{align}
This derivation demonstrates that the overhead of DIPrune scales linearly with the visual token count $N_v$ and the query window size $L_{win}$, rather than the quadratic complexity $O(N_v^2)$ associated with full self-attention. This efficiency ensures that DIPrune imposes minimal latency while enabling substantial acceleration in subsequent layers.

\subsection{Robustness of Query Window Strategy}
\label{sec:window_ablation}

To further validate the claim that the text query windowing strategy achieves performance comparable 
to full-text attention, we evaluate DIPrune across query-window ratios ranging from 0.3 to 1.0 
on LLaVA-v1.5-7B. As shown in Table~\ref{tab:window_ablation}, performance saturates around a 
ratio of 0.8, which is therefore adopted as our default setting.

\begin{table}[H]
\centering
\caption{Performance under different query-window ratios on LLaVA-v1.5-7B.}
\label{tab:window_ablation}
\setlength{\tabcolsep}{6pt}
\renewcommand{\arraystretch}{1.1}
\begin{tabular}{lcccccc}
\toprule
Window Ratio & 0.3 & 0.5 & 0.7 & \textbf{0.8} & 0.9 & 1.0 \\
\midrule
TextVQA & 56.11 & 56.83 & 57.11 & \textbf{57.21} & 57.13 & 57.19 \\
GQA     & 57.74 & 57.74 & 58.01 & \textbf{58.88} & 58.68 & 58.72 \\
\bottomrule
\end{tabular}
\end{table}

Furthermore, comparing the default ratio ($L_{win}/L_{total} = 0.8$) against the full-attention 
baseline ($L_{win}/L_{total} = 1.0$) across different prompt lengths on TextVQA, the performance 
gap is $-0.15$ / $+0.44$ / $-0.18$ for short, medium, and long prompts respectively, with an 
overall difference of $+0.06$. These results confirm that the windowing strategy maintains robust 
performance across a wide range of ratios and prompt lengths, validating its use as an efficient 
approximation without sacrificing accuracy.

\subsection{Implementation Details of Decoupled Attention}
\label{supp:attn_detail}

In this section, we provide a detailed description of the implementation specifications 
of the decoupled attention computation.

\noindent\textbf{Detailed Implementation.}
At the pruning layer $l$, the hidden states are projected onto queries and keys. 
We extract the visual key block $K_{\text{vis}} \in \mathbb{R}^{N_v \times D}$ and the recent 
text query window $Q_{\text{sub}} \in \mathbb{R}^{L_{\text{win}} \times D}$, and compute 
per-head attention as follows:
\begin{equation}
    A^{(h)} = \text{softmax}\!\left(\frac{Q_{\text{sub}}^{(h)} \left(K_{\text{vis}}^{(h)}\right)^\top}{\sqrt{d_h}}\right), \quad h = 1, \ldots, H
\end{equation}
where $d_h = D / H$ is the per-head dimension. The per-head attention maps are then aggregated 
via mean pooling across all $H$ heads to obtain a unified attention map:
\begin{equation}
    \bar{A} = \frac{1}{H} \sum_{h=1}^{H} A^{(h)} \in \mathbb{R}^{L_{\text{win}} \times N_v}
\end{equation}
The raw saliency score for each visual token $i$ is obtained by averaging over all query positions 
in the window:
\begin{equation}
    a_i = \frac{1}{L_{\text{win}}} \sum_{j=1}^{L_{\text{win}}} \bar{A}[j, i], \quad i = 1, \ldots, N_v
\end{equation}
This yields a scalar importance score $a_i$ for each visual token, reflecting its overall 
relevance to the text queries within the current window. The final dual importance score 
is then computed following Eq.~\ref{eq:align_importance} and Eq.~\ref{eq:s_evol} in the main paper, identical to the standard 
DIPrune scoring pipeline.

\section{Additional Experiments and Analysis}
\subsection{More Experimental Validation on Generalization and Scalability}
\label{supp:model_Generalization}

\noindent\textbf{Model Scaling to Larger Models: LLaVA-v1.5-13B.} To assess cross-scale generalization, we evaluate DIPrune on LLaVA-v1.5-13B across various pruning intensities. As shown in Table~\ref{tab:13B-performance_comparison}, DIPrune demonstrates exceptional resilience on the larger 13B architecture. Specifically, under aggressive compression (128 tokens, $\downarrow 77.8\%$), our method achieves $99.9\%$ average performance retention compared to the unpruned upper bound. This significantly outperforms the $99.5\%$ retention observed on the 7B counterpart under the same budget, delivering consistently positive gains across nearly all competency dimensions. Even under extreme compression conditions (retaining only 64 tokens, $\downarrow 88.9\%$), DIPrune maintains $98.9\%$ performance, substantially leading the best-performing baseline method, DART, by $+2.8\%$. Overall, these results demonstrate the robust cross-scale transferability of DIPrune and support a plausible scale-dynamics synergy interpretation: larger-capacity models may exhibit more pronounced inter-layer attention reallocation, rendering the rank-displacement-based $\Delta$Rank trend signal more discriminative. This enables DIPrune to more effectively preserve emergent evidence that is critical for later-stage reasoning under tight token budgets.

\begin{table}[!t]
\centering
\caption{{Performance Comparison on LLaVA-v1.5-13B across Diverse Benchmarks}. Results are reported under different pruning ratios. Gray shading denotes our method, and \textbf{bold} indicates the best results.}
\label{tab:13B-performance_comparison}
\small
\begin{tabular}{@{}lrrrrrrrr@{}}
\toprule
\textbf{Methods} & \textbf{GQA} & \textbf{MMB} & \textbf{MMB\textsubscript{CN}} & \textbf{MME} & \textbf{POPE} & \textbf{SQA} & \textbf{VQA\textsubscript{Text}} & \textbf{Average} \\
\midrule\midrule

Upper Bound, 576 Tokens & 63.3 & 68.9 & 62.3 & 1818 & 85.9 & 72.8 & 61.3 & 100\% \\
\midrule

\multicolumn{9}{@{}l}{\textbf{LLaVA-1.5 13B \hfill Retain 192 Tokens ($\downarrow 66.7\%$)}} \\
\cmidrule(r){1-9}
FastV (ECCV'24)               & 59.1 & 54.0 & 51.2 & 1641 & 82.3 & 56.4 & 51.6 & 86.0\%  \\
VisionZip (CVPR'25)           & 59.1 & 66.9 & - & 1754 & 85.1 & 73.5 & 59.5 & 97.3\%  \\

SparseVLM (ICML'25)           & 58.7 & 67.4 & 61.0 & 1768 & 82.2 & 73.1 & 55.4 & 96.0\%  \\
DART (EMNLP'25)               & 62.1 & 68.2 & 61.4 & \textbf{1855} & 84.0 & 73.6 & 60.2
 & 99.3\%

\\
\rowcolor{gray!15}DIPrune(Ours)                 & \textbf{62.7} & \textbf{68.4} & \textbf{63.4} & 1817 & \textbf{87.0} & 72.8 & \textbf{61.1} & \textbf{100.1\%}  \\

\midrule

\multicolumn{9}{@{}l}{\textbf{LLaVA-1.5 13B \hfill Retain 128 Tokens ($\downarrow 77.8\%$)}} \\
\cmidrule(r){1-9}
FastV (ECCV'24)               & 57.7 & 57.9 & 48.8 & 1673 & 79.3 & 57.0 & 56.0 & 86.8\%  \\
VisionZip (CVPR'25)           & 57.9 & 66.7 & - & 1743 & 85.2 & 74.0 & 58.7 & 96.8\%  \\

SparseVLM (ICML'25)           & 57.9 & 65.8 & 55.8 & 1774 & 81.1 & 69.9 & 49.9 & 92.3\%  \\
DART (EMNLP'25)               & 60.9 & 67.4 & 60.7 & \textbf{1839} & 81.8 & \textbf{74.3} & 59.0
 & 98.0\%
\\
\rowcolor{gray!15}DIPrune(Ours)                 & \textbf{62.3} & \textbf{67.5} & \textbf{63.8} & 1815 & \textbf{87.0} & 72.7 & \textbf{60.8} & \textbf{99.9\%}  \\

\midrule

\multicolumn{9}{@{}l}{\textbf{LLaVA-1.5 13B \hfill Retain 64 Tokens ($\downarrow 88.9\%$)}} \\
\cmidrule(r){1-9}
FastV (ECCV'24)               & 53.7 & 50.9 & 42.1 & 1567 & 69.3 & 56.8 & 47.1& 78.3\%  \\
VisionZip (CVPR'25)           & 56.2 & 64.9 & - & 1676 & 76.0 & \textbf{74.4} & 57.4 & 93.3\%  \\
DivPrune (CVPR'25)           & 57.9 & 64.3 & - & 1779 & 84.7 & 72.4 & 58.7 & 96.1\%  \\
SparseVLM (ICML'25)           & 50.6 & 61.3 & 54.8 & 1502 & 65.0 & 69.0 & 47.2 & 83.9\%  \\
DART (EMNLP'25)               & 57.1 & 65.4 & 59.3 & 1722 & 75.4 & 74.1 & 55.9
 & 96.1\%

\\
\rowcolor{gray!15}DIPrune(Ours)                 & \textbf{60.5} & \textbf{66.8} & \textbf{63.5} & \textbf{1795} & \textbf{87.3} & 72.9 & \textbf{59.5} & \textbf{98.9\%}  \\

\bottomrule
\end{tabular}
\end{table}

\noindent\textbf{Cross-Modality Generalization: Video-LLaVA-7B.} We further apply DIPrune to the video understanding model Video-LLaVA-7B to assess its applicability to spatiotemporal token sequences. Compared to static images, video inputs typically exhibit stronger temporal redundancy and more complex dynamic disturbances. As shown in Table~\ref{tab:video_qa}, experimental results demonstrate that DIPrune substantially reduces the token budget while maintaining superior overall performance, demonstrating strong cross-modality generalization.

\begin{table}[!t]
\centering
\caption{Video QA Evaluations on Video-LLaVA-7B under 50\% Token Retention.}
\label{tab:video_qa}

\small 
\setlength{\tabcolsep}{6pt} 
\renewcommand{\arraystretch}{1.2} 

\begin{tabular}{@{}lcccccc@{}}
\toprule
\multirow{2}{*}{\textbf{Methods}} & \multicolumn{2}{c}{\textbf{MSVD-QA}} & \multicolumn{2}{c}{\textbf{MSRVTT-QA}} & \multicolumn{2}{c}{\textbf{Average}} \\
\cmidrule(lr){2-3} \cmidrule(lr){4-5} \cmidrule(lr){6-7}
 & \textbf{Acc.} & \textbf{Score} & \textbf{Acc.} & \textbf{Score} & \textbf{Acc.} & \textbf{Score} \\
\midrule\midrule

Upper Bound (Video-LLaVA) & 70.2 & 3.9 & 57.3 & 3.5 & 63.8 & 3.7 \\
\midrule

\multicolumn{7}{@{}l}{\textbf{Retain 50\% Tokens ($\downarrow 50\%$)}} \\
\cmidrule(r){1-7}
FastV (ECCV'24) & 71.0 & 3.9 & 55.0 & 3.5 & 63.0 & 3.7 \\
FasterVLM (arXiv'24) & 70.5 & 3.9 & 56.2 & 3.5 & 63.4 & 3.7 \\
DART (EMNLP'25) & 71.0 & 4.0 & 56.7 & 3.6 & 63.8 & 3.8 \\
HoloV (NeurIPS'25) & 71.0 & 4.0 & 56.5 & 3.6 & 63.7 & 3.8 \\

\rowcolor{gray!15} DIPrune (Ours) & \textbf{71.3} & \textbf{4.0} & \textbf{56.9} & \textbf{3.6} & \textbf{64.1} & \textbf{3.8} \\
\bottomrule
\end{tabular}
\end{table}

\noindent\textbf{Fine-Grained Task Generalization: Qwen2.5-VL-7B.} To further verify the efficacy of DIPrune on tasks requiring precise spatial reasoning, we conduct experiments on the challenging visual grounding benchmarks: RefCOCO, RefCOCO+, and RefCOCOg. Unlike general VQA tasks that rely heavily on global semantic abstraction, visual grounding demands the retention of fine-grained spatial features to accurately regress bounding box coordinates for specific textual descriptions. As illustrated in Table~\ref{tab:refcoco_comparison}, DIPrune consistently achieves superior performance compared to baseline methods. This success highlights a key advantage of our framework: the Semantic Alignment ($S_{align}$) component effectively anchors tokens corresponding to the textual referring expressions (e.g., ``the cat on the left''), while the Semantic Evolution ($S_{evol}$) component ensures the preservation of tokens critical for subsequent coordinate regression, particularly those that emerge only in deeper layers. This demonstrates that DIPrune does not merely retain salient objects but actively preserves the spatial-semantic integrity necessary for dense localization tasks.

\begin{table}[H]
\centering
\caption{{Performance Comparisons on Qwen2.5-VL-7B across Three Referring Expression Comprehension Benchmarks.} Results are reported under the aggressive pruning ratio (retaining 25\% tokens). Gray shading denotes our method, and \textbf{bold} indicates the best results.}
\label{tab:refcoco_comparison}

\small 
\setlength{\tabcolsep}{4.2pt}
\renewcommand{\arraystretch}{1.2}

\begin{tabular}{@{}lccccccccc@{}}
\toprule
\multirow{2}{*}{\textbf{Method}} & \multicolumn{3}{c}{\textbf{RefCOCO}} & \multicolumn{3}{c}{\textbf{RefCOCO+}} & \multicolumn{2}{c}{\textbf{RefCOCOg}} & \multirow{2}{*}{\textbf{Average}} \\
\cmidrule(lr){2-4} \cmidrule(lr){5-7} \cmidrule(lr){8-9}
 & \textbf{val} & \textbf{testA} & \textbf{testB} & \textbf{val} & \textbf{testA} & \textbf{testB} & \textbf{val} & \textbf{test} & \\
\midrule\midrule

Upper Bound (Qwen2.5-VL-7B) & 89.45 & 92.56 & 85.16 & 83.50 & 89.02 & 79.15 & 86.76 & 87.24 & 100\% \\
\midrule

\multicolumn{10}{@{}l}{\textbf{Qwen2.5-VL-7B \hfill Retain 25\% Tokens ($\downarrow 75\%$)}} \\
\cmidrule(r){1-10}

FastV (ECCV'24) & 43.57 & 46.81 & 40.86 & 39.47 & 43.78 & 36.02 & 43.04 & 42.69 & 48.5\% \\
PyramidDrop (CVPR'25) & 46.46 & 53.83 & 37.23 & 42.29 & 47.76 & 32.81 & 45.32 & 44.91 & 50.4\% \\
VScan (TMLR) & 74.32 & 79.05 & 68.22 & 67.22 & 73.72 & \textbf{58.95} & 69.42 & 69.43 & 80.7\% \\

\rowcolor{gray!15} DIPrune (Ours) & \textbf{74.88} & \textbf{79.76} & \textbf{68.26} & \textbf{68.16} & \textbf{74.54} & 58.80 & \textbf{73.47} & \textbf{73.27} & \textbf{82.3\%} \\

\bottomrule
\end{tabular}
\end{table}

\noindent\textbf{Cross-Architecture Scalability: Qwen3-VL Series.}
To further demonstrate the scalability and architecture-agnostic nature of DIPrune, we extend 
our evaluation to two additional models from the Qwen3-VL series: Qwen3-VL-8B (Dense) and 
Qwen3-VL-30B-A3B (MoE). These models represent distinct architectural paradigms---a standard 
dense transformer and a Mixture-of-Experts (MoE) architecture---and are evaluated against the 
recent strong baseline HoloV across three pruning ratios. Results are reported on MME, MMBench, 
and POPE in Table~\ref{tab:qwen3vl}.

\begin{table}[H]
\centering
\caption{Performance comparison on Qwen3-VL series under different pruning ratios 
(MME / MMBench / POPE). Gray shading denotes our method, and \textbf{bold} indicates 
the best results.}
\label{tab:qwen3vl}
\setlength{\tabcolsep}{5pt}
\renewcommand{\arraystretch}{1.15}
\resizebox{\linewidth}{!}{%
\begin{tabular}{llccc}
\toprule
\textbf{Model} & \textbf{Method} & \textbf{66.7\%} & \textbf{77.8\%} & \textbf{88.9\%} \\
\midrule
\multirow{2}{*}{Qwen3-VL-8B (Dense)}
& HoloV
& 2007 / 78.6 / 83.5
& 1956 / 76.8 / 80.5
& 1798 / 67.7 / 74.6 \\
& \cellcolor{gray!15}DIPrune (Ours)
& \cellcolor{gray!15}\textbf{2152 / 81.3 / 88.3}
& \cellcolor{gray!15}\textbf{2029 / 78.6 / 87.8}
& \cellcolor{gray!15}\textbf{1918 / 75.8 / 85.9} \\
\midrule
\multirow{2}{*}{Qwen3-VL-30B-A3B (MoE)}
& HoloV
& 2076 / 80.4 / 84.5
& 1941 / 76.3 / 81.2
& 1724 / 67.1 / 71.7 \\
& \cellcolor{gray!15}DIPrune (Ours)
& \cellcolor{gray!15}\textbf{2278 / 85.1 / 89.1}
& \cellcolor{gray!15}\textbf{2089 / 83.8 / 86.9}
& \cellcolor{gray!15}\textbf{1877 / 78.8 / 75.2} \\
\bottomrule
\end{tabular}%
}
\end{table}

As shown in Table~\ref{tab:qwen3vl}, DIPrune consistently outperforms HoloV across all 
pruning ratios on both models. Notably, the performance advantage becomes more pronounced 
under aggressive compression (88.9\% pruning ratio), where DIPrune surpasses HoloV by 
up to \textbf{+11.3} on POPE for Qwen3-VL-8B and \textbf{+11.7} on MMBench for Qwen3-VL-30B-A3B. 
Furthermore, the consistent gains observed from the 8B dense model to the 30B-A3B MoE 
model confirm that DIPrune is not restricted to specific backbone design, and generalizes 
effectively across both model scales and architectural variants of the latest generation MLLMs.

\textbf{Generative Task Generalization.} We further evaluate DIPrune on detailed captioning
(DetailCaps~\citep{dong2024benchmarking}), open-ended hallucination (CHAIR~\citep{rohrbach2018object}), multi-turn reasoning (ConvBench~\citep{liu2024convbench}), and fine-grained perception
(OCRBench~\citep{liu2024ocrbench}, ChartQA~\citep{masry2022chartqa}). As shown in Tab.~\ref{tab:generative}, DIPrune consistently achieves superior performance
across both models and pruning ratios, with the captioning gain growing steadily as the pruning ratio increases,
confirming its effectiveness beyond discriminative tasks.

\begin{table}[H]

\centering
\footnotesize
\setlength{\tabcolsep}{3pt}
\caption{Results on generative and fine-grained benchmarks. Cap.: DetailCaps (CAPTURE); CHAIR: CHAIR$_i$;
Conv.: ConvBench; OCR: OCRBench; Chart: ChartQA (\%).}
\label{tab:generative}
\vspace{-6pt}
\begin{tabular*}{\linewidth}{@{\extracolsep{\fill}} ll ccccc ccccc @{}}
\toprule
\multirow{2}{*}{Ratio} & \multirow{2}{*}{Method} & \multicolumn{5}{c}{LLaVA-1.5-7B} & \multicolumn{5}{c}{Qwen2.5-VL-7B} \\
\cmidrule(lr){3-7}\cmidrule(lr){8-12}
 & & Cap.$\uparrow$ & CHAIR$\downarrow$ & Conv.$\uparrow$ & OCR$\uparrow$ & Chart$\uparrow$
   & Cap.$\uparrow$ & CHAIR$\downarrow$ & Conv.$\uparrow$ & OCR$\uparrow$ & Chart$\uparrow$ \\
\midrule
\multirow{3}{*}{66.7\%} & FastV & .496 & 13.45 & 4.27 & 300 & 17.4 & .543 & 8.68 & 4.31 & 426 & 63.4 \\
 & HoloV & .489 & 13.55 & 4.29 & 299 & 15.4 & .599 & 10.13 & 6.32 & 481 & 57.6 \\
\rowcolor{gray!15}
\cellcolor{white} & DIPrune & \textbf{.501} & \textbf{12.91} & \textbf{4.32} & \textbf{307} & \textbf{17.5} & \textbf{.613} & \textbf{8.66} & \textbf{6.40} & \textbf{492} & \textbf{71.8} \\
\midrule
\multirow{3}{*}{77.8\%} & FastV & .485 & 13.20 & 4.07 & 287 & 16.5 & .520 & \textbf{8.23} & 4.24 & 332 & 57.8 \\
 & HoloV & .475 & 14.00 & 4.16 & 291 & 17.0 & .581 & 10.12 & 6.21 & 360 & 44.7 \\
\rowcolor{gray!15}
\cellcolor{white} & DIPrune & \textbf{.493} & \textbf{11.54} & \textbf{4.28} & \textbf{293} & \textbf{17.5} & \textbf{.604} & 8.68 & \textbf{6.25} & \textbf{378} & \textbf{63.4} \\
\midrule
\multirow{3}{*}{88.9\%} & FastV & .455 & 17.51 & 3.77 & 236 & 14.8 & .446 & 9.68 & 3.80 & 178 & 41.8 \\
 & HoloV & .453 & 14.98 & 4.08 & 247 & 16.2 & .552 & 10.36 & 5.76 & 218 & 30.4 \\
\rowcolor{gray!15}
\cellcolor{white} & DIPrune & \textbf{.482} & \textbf{14.79} & \textbf{4.16} & \textbf{270} & \textbf{16.4} & \textbf{.582} & \textbf{8.12} & \textbf{5.82} & \textbf{220} & \textbf{42.9} \\
\bottomrule
\end{tabular*}
\end{table}

\textbf{Stratified Analysis on Query Length and Scene Density.}\label{app:challenge} To further quantify the limitations discussed
in Sec.~\ref{sec:conclusion}, we stratify GQA by query length and the above CHAIR results by the number of
ground-truth instances on LLaVA-1.5-7B (88.9\% pruning; Tab.~\ref{tab:challenge}). DIPrune leads HoloV across
all query lengths, while the margin shrinks for the shortest queries (+2.0\% at 3--5 words vs.\ +4.1\% at
6--11 words), consistent with a weaker alignment signal. On CHAIR$_i$, DIPrune performs best in sparse and
dense scenes, whereas HoloV is slightly better in medium-density scenes, indicating that cluttered scenes
remain a challenging regime.
\begin{table}[H]
\centering
\footnotesize
\setlength{\tabcolsep}{3pt}
\caption{Stratified analysis on LLaVA-1.5-7B (88.9\% pruning). (a) GQA accuracy (\%) by query length;
(b) CHAIR$_i$ ($\downarrow$) by the number of ground-truth instances. Bold: best among pruning methods.}
\label{tab:challenge}
\vspace{-6pt}
\begin{minipage}[t]{0.49\linewidth}
\centering
(a) Query length\\[2pt]
\begin{tabular*}{\linewidth}{@{\extracolsep{\fill}} lccccc @{}}
\toprule
Length & \#Samples & Full & FastV & HoloV & \cellcolor{gray!15}DIPrune \\
\midrule
3--5   & 2,071  & 51.1 & 41.7 & 46.2 & \cellcolor{gray!15}\textbf{48.2} \\
6--11  & 8,425  & 62.9 & 44.9 & 56.0 & \cellcolor{gray!15}\textbf{60.1} \\
12--25 & 2,082  & 68.8 & 55.3 & 60.3 & \cellcolor{gray!15}\textbf{64.7} \\
\midrule
All    & 12,578 & 61.9 & 46.1 & 55.1 & \cellcolor{gray!15}\textbf{58.9} \\
\bottomrule
\end{tabular*}
\end{minipage}\hfill
\begin{minipage}[t]{0.49\linewidth}
\centering
(b) Scene density\\[2pt]
\begin{tabular*}{\linewidth}{@{\extracolsep{\fill}} lccccc @{}}
\toprule
\#Inst. & \#Images & Full & FastV & HoloV & \cellcolor{gray!15}DIPrune \\
\midrule
1--2       & 217 & 14.41 & 20.83 & 16.38 & \cellcolor{gray!15}\textbf{14.91} \\
3--4       & 187 & 13.14 & 14.09 & \textbf{12.76} & \cellcolor{gray!15}14.06 \\
$\geq$5    & 93  & 14.44 & 16.64 & 16.17 & \cellcolor{gray!15}\textbf{15.98} \\
\midrule
All        & 497 & 13.94 & 17.51 & 14.98 & \cellcolor{gray!15}\textbf{14.79} \\
\bottomrule
\end{tabular*}
\end{minipage}
\end{table}

\subsection{More Ablation Analysis}
\label{supp:model_ablation}

\textbf{Empirical Validation of Proxy Scores.}\label{app:proxy}
As $S_{\text{align}}$ and $S_{\text{evol}}$ are tractable proxies of $\mathcal{I}_{\text{Intra}}$ and
$\mathcal{I}_{\text{Inter}}$, we validate them on LLaVA-1.5-7B (64 tokens; Tab.~\ref{tab:proxy}).
(a) Since the attention logit couples projected feature saliency with query-directional alignment,
$S_{\text{align}}$ acts as a task-conditioned instantiation of $\mathcal{I}_{\text{Intra}}$: replacing it with
the raw feature magnitude $\|v_i\|^2$ consistently degrades performance, with or without $S_{\text{evol}}$.
(b) Using the gradient energy $G_i=\|\partial\mathcal{L}_{\text{ans}}/\partial h_i^{(10)}\|_2^2$ of the unpruned model
as an offline reference, we report GradMass, i.e., the fraction of $\sum_i G_i$ captured by the retained tokens.
$S_{\text{evol}}$ alone achieves the highest GradMass (61.8\% vs.\ 16.0--16.5\%), supporting it as an effective
proxy for $\mathcal{I}_{\text{Inter}}$, while combining both terms yields the best accuracy, confirming their
complementarity.

\begin{table}[H]
\centering
\footnotesize
\caption{Validation of proxy scores on LLaVA-1.5-7B (64 tokens). (a) Intra-layer instantiation;
(b) GradMass: fraction of gradient energy captured by retained tokens.}
\label{tab:proxy}
\vspace{-6pt}
\begin{minipage}[t]{0.48\linewidth}
\centering
\textit{(a) Intra-layer term}\\[2pt]
\begin{tabular*}{\linewidth}{@{\extracolsep{\fill}} lcc @{}}
\toprule
Configuration & GQA$\uparrow$ & MME$\uparrow$ \\
\midrule
$\|v_i\|^2$ only & 53.4 & 1698 \\
$\|v_i\|^2$ + $S_{\text{evol}}$ & 55.7 & 1763 \\
$S_{\text{align}}$ only & 58.1 & 1722 \\
\rowcolor{gray!15}
DIPrune & \textbf{58.9} & \textbf{1807} \\
\bottomrule
\end{tabular*}
\end{minipage}\hfill
\begin{minipage}[t]{0.48\linewidth}
\centering
\textit{(b) Inter-layer term}\\[2pt]
\begin{tabular*}{\linewidth}{@{\extracolsep{\fill}} lcc @{}}
\toprule
Method & GradMass$\uparrow$ & GQA$\uparrow$ \\
\midrule
HoloV & 16.5\% & 55.1 \\
$S_{\text{align}}$ only & 16.0\% & 58.1 \\
$S_{\text{evol}}$ only & \textbf{61.8\%} & 57.7 \\
\rowcolor{gray!15}
DIPrune & 59.7\% & \textbf{58.9} \\
\bottomrule
\end{tabular*}
\end{minipage}
\end{table}

\noindent\textbf{Impact of Rank Displacement Step \texorpdfstring{$\delta$}{delta}.} We evaluate the rank displacement step $\delta \in \{1, 2, 3, 4\}$ to determine the optimal observation scope. As shown in Table~\ref{tab:ablation_step},  LLaVA-NeXT yield optimal performance at \textbf{$\delta=1$} and degrade as the window widens. This observation suggests that critical evolutionary signals are likely derived from adjacent layers, whereas wider windows may introduce irrelevant historical noise. Consequently, we fix $\delta=1$ as the default setting to maximize capture precision with minimal memory overhead.

\noindent\textbf{Impact of Rank Displacement Clipping Threshold ($M_{max}$).} The clipping threshold $M_{max}$ in Eq.~\ref{eq:s_evol} is designed to balance the sensitivity toward semantic emergence against robustness to stochastic fluctuations. As summarized in Table~\ref{tab:ablation_m_max}, varying $M_{max} \in \{50, 100, 200, 300, 400\}$ on LLaVA-NeXT-7B reveals three distinct behavioral regimes. Initially, in the under-sensitivity regime ($M_{max} \le 100$), restrictive thresholds constrain performance on fine-grained tasks such as TextVQA (58.3\%). The model achieves a balanced trade-off at $M_{max}=200$, reaching superior performance on TextVQA (58.7\%) while maintaining high stability on GQA and SQA. This reflects that the threshold effectively encapsulates the typical magnitude of rank displacement for valid emerging tokens, allowing the DIPrune framework to accurately anchor information that becomes critical only in deeper layers of reasoning. However, further increasing the threshold into the noise interference regime ($M_{max} \ge 300$) leads to performance degradation on reasoning-intensive benchmarks like GQA, even as generalized perception metrics such as MME continue to scale (reaching 1807). We hypothesize that excessively high thresholds inadvertently incorporate non-semantic-related noise. Consequently, we adopt $M_{max} = 200$ as the default configuration to strike a favorable trade-off between capturing semantic emergence and suppressing non-stationary noise.

\begin{table}[H]
    \centering
    \begin{minipage}{0.45\textwidth}
        \centering
        \caption{Impact of $\delta$ on LLaVA-NeXT-7B. $\delta=1$ is the default setting.}
        \label{tab:ablation_step}
        \small 
        \renewcommand{\arraystretch}{1.2}
        \setlength{\tabcolsep}{8pt} 
        \begin{tabular}{@{}lcccc@{}}
            \toprule
            \multirow{2}{*}{\textbf{Metric}} & \multicolumn{4}{c}{\textbf{Step ($\delta$)}} \\
            \cmidrule(l){2-5}
             & 1 & 2 & 3 & 4 \\
            \midrule
            MME  & \textbf{1788} & 1767 & 1782 & 1780 \\
            GQA  & \textbf{63.1} & 62.8 & 62.2 & 62.1 \\
            SQA  & \textbf{69.3} & 69.1 & 69.2 & 69.2 \\
            POPE & \textbf{87.7} & 87.3 & 87.4 & 87.2 \\
            \bottomrule
        \end{tabular}
    \end{minipage}
    \hfill 
    \begin{minipage}{0.52\textwidth}
        \centering
        \caption{Ablation Study on $M_{max}$ on LLaVA-NeXT-7B. $M_{max}=200$ is the default setting.}
        \label{tab:ablation_m_max}
        \small
        \renewcommand{\arraystretch}{1.2}
        \setlength{\tabcolsep}{6pt}
        \begin{tabular}{@{}lccccc@{}}
            \toprule
            \multirow{2}{*}{\textbf{Metric}} & \multicolumn{5}{c}{\textbf{Clipping Threshold ($M_{max}$)}} \\
            \cmidrule(l){2-6}
             & 50 & 100 & 200 & 300 & 400 \\
            \midrule
            TextVQA & 58.3 & 58.3 & 58.7 & \textbf{59.3} & 59.3 \\
            GQA     & 63.1 & 63.1 &  \textbf{63.1} & 63.1 & 62.6 \\
            SQA     & \textbf{69.5} & 69.5 & 69.3 & 69.1 & 69.0 \\
            MME     & 1785 & 1780 & 1788 & 1796 & \textbf{1807} \\
            \bottomrule
        \end{tabular}
    \end{minipage}
\end{table}

\noindent\textbf{Extended Sensitivity Analysis on Dual Importance Weights.}
To evaluate the generalizability of the proposed dual importance scoring mechanism, we extend the sensitivity analysis of hyperparameters $\alpha$ and $\beta$ to the Qwen2.5-VL-7B architecture. Following the default experimental protocol, we perform one-shot pruning at the 10th transformer layer  and evaluate the performance trajectory across a spectrum of fusion ratios $\alpha:\beta$ under the constraint $\alpha + \beta = 1$. 

As shown in Table~\ref{tab:sensitivity_qwen}, the empirical results exhibit a consistent peak performance at $\alpha=0.6$ and $\beta=0.4$. This persistent optimality across diverse architectures suggests that DIPrune effectively captures architectural invariants within MLLMs. By maintaining a precise equilibrium between dynamic semantic evolving and stable saliency anchoring, our framework minimizes the cumulative approximation error induced by pruning while substantially reducing the tuning overhead for future MLLM deployments.

\begin{table}[H]
\centering
\caption{Extended Sensitivity Analysis on Qwen2.5-VL-7B. We evaluate the performance trajectory across a spectrum of fusion ratios $\alpha : \beta$. The default setting ($\alpha=0.6, \beta=0.4$) consistently achieves high performance. Underlined ratios correspond to using only $S_{align}$ or $S_{evol}$.}
\label{tab:sensitivity_qwen}

\small
\setlength{\tabcolsep}{10pt}
\renewcommand{\arraystretch}{1.2}

\begin{tabular}{@{}lccccccc@{}}
\toprule
\textbf{Metrics} & \underline{1.0 : 0.0} & 0.8 : 0.2 & \textbf{0.6 : 0.4} & 0.5 : 0.5 & 0.4 : 0.6 & 0.2 : 0.8 & \underline{0.0 : 1.0} \\
\midrule
MME & 1943 & 2046 & \textbf{2193} & 2134 & 2097 & 2019 & 1984 \\
MMB & 79.3 & 80.7 & \textbf{81.4} & 81.2 & 80.9 & 80.6 & 78.8 \\
POPE & 84.4 & 85.9 & \textbf{87.5} & 86.4 & 86.2 & 85.6 & 83.7 \\
\bottomrule
\end{tabular}
\end{table}

\noindent\textbf{Analysis on Pruning Layer Selection.}
The ablation study in Figure~\ref{fig:ablation_layer_span} demonstrates that the 10th layer serves as the optimal pruning 
point for LLaVA-NeXT-7B. Here we provide a deeper discussion on whether and how this choice 
generalizes across models of different depths, visual token counts, and task types.

We do not treat the 10th layer as a universal constant. Rather, the optimal pruning point 
reflects a fundamental trade-off: the model must accumulate sufficient semantic information 
for reliable token ranking, while retaining enough remaining layers for pruning to yield 
meaningful efficiency gains. This trade-off is precisely captured by our layer-wise ablation 
(Figure~\ref{fig:ablation_layer_span}): pruning too early degrades accuracy because multimodal representations are not 
yet mature, whereas pruning too late offers diminishing accuracy improvements at the cost of 
reduced computational savings.

More generally, the optimal pruning layer depends on three interacting factors:
\begin{itemize}
    \item \textbf{Model depth.} Deeper models naturally develop stable representations at later absolute layers; 
yet the relative pruning point, expressed as a fraction of total depth, tends to 
remain consistent across architectures.

    \item \textbf{Visual token count.} Models processing denser visual token sequences 
    (e.g., LLaVA-NeXT with 2880 tokens) benefit more from earlier pruning to amortize 
    the quadratic attention cost, whereas models with fewer tokens can afford to prune 
    slightly later without significant latency overhead.
    \item \textbf{Task type.} Tasks requiring fine-grained spatial reasoning (e.g., visual 
    grounding) may favor deeper pruning to preserve spatially precise token signals, 
    while general VQA tasks are more tolerant of earlier pruning.
\end{itemize}

In practice, rather than relying on a fixed absolute layer index, we recommend selecting 
the pruning point based on the relative stage of representation maturity---typically 
within the early-to-middle portion of the LLM backbone. For a given model, an effective 
operating point can be efficiently identified by scanning a small set of candidate layers 
on a lightweight validation subset, which incurs negligible additional cost. This procedure 
is consistent with standard hyperparameter selection in training-free methods and requires 
no gradient computation or model fine-tuning.

\noindent\textbf{Performance Across Extended Token Budgets.}
The performance advantage of DIPrune over competing methods becomes increasingly pronounced 
as the token budget decreases. This trend stems from its core design: under generous budgets, 
most informative tokens are commonly retained across methods, leaving limited room for 
differentiation. Under high compression ratios, however, the ability to effectively identify 
and preserve tokens with semantic evolutionary potential becomes the decisive factor. By 
jointly modeling intra-layer semantic alignment and inter-layer semantic evolution, DIPrune 
demonstrates a more substantial competitive advantage precisely under resource-constrained, 
extreme compression settings.
To further substantiate this, we report results across four additional token budgets on 
LLaVA-v1.5-7B, presented as TextVQA / GQA / MME in Table~\ref{tab:extended_budget}.

\begin{table}[H]
\centering
\caption{Performance comparison across extended token budgets on LLaVA-v1.5-7B 
(TextVQA / GQA / MME). \textbf{Bold} indicates the best results.}
\label{tab:extended_budget}
\setlength{\tabcolsep}{5pt}
\renewcommand{\arraystretch}{1.15}
\resizebox{\linewidth}{!}{%
\begin{tabular}{lcccc}
\toprule
\textbf{Method} & \textbf{432 Tokens} & \textbf{288 Tokens} & \textbf{32 Tokens} & \textbf{16 Tokens} \\
\midrule
VisionZip  & 58.08 / 61.57 / 1836 & 57.92 / 60.32 / 1811 & 53.03 / 52.15 / 1580 & 49.76 / 46.95 / 1350 \\
HoloV      & 52.24 / 55.83 / 1605 & 54.63 / 57.83 / 1620 & 53.55 / 52.37 / 1596 & 50.40 / 48.60 / 1430 \\
\rowcolor{gray!15}\textbf{DIPrune (Ours)} & \textbf{58.32 / 62.00 / 1862} & \textbf{58.25 / 61.93 / 1828} & \textbf{55.96 / 56.20 / 1756} & \textbf{53.84 / 54.60 / 1712} \\
\bottomrule
\end{tabular}%
}
\end{table}

As shown in Table~\ref{tab:extended_budget}, DIPrune achieves the best results across all 
four token budgets. Notably, the performance gap widens progressively as the budget decreases: 
under the most aggressive setting (16 tokens), DIPrune outperforms VisionZip by \textbf{+4.08} on 
TextVQA and \textbf{+7.65} on GQA, and surpasses HoloV by \textbf{+3.44} on TextVQA and \textbf{+6.00} on GQA. 
These results confirm that the joint modeling of intra-layer semantic alignment and inter-layer 
semantic evolution confers increasing advantages under high compression ratios, where 
identifying semantically evolving tokens is most critical.

\subsection{More Efficiency Analysis}
\label{app:eff}

\textbf{Setup.} All efficiency experiments are conducted on a single NVIDIA H800 80GB GPU with FP16
precision, CUDA 12.1, PyTorch 2.1.2, Transformers 4.37.2, and batch size 1. For TTFT/ITT
(Tab.~\ref{tab:efficiency} and Tab.~\ref{tab:eff_models}), GPU-side latency is measured with
\texttt{torch.cuda.Event} over 500 POPE samples (after 5 warm-up samples) with a fixed generation length
of 32 tokens. Unless otherwise stated, baselines follow their default pruning configurations.

\textbf{Cross-Architecture Latency and Overhead.} Tab.~\ref{tab:eff_models} extends
Tab.~\ref{tab:efficiency}(a) to LLaVA-NeXT-7B and Qwen2.5-VL-7B. DIPrune reduces TTFT by
55.6\%/56.9\%/35.8\% on LLaVA-1.5/LLaVA-NeXT/Qwen2.5-VL while ITT remains nearly identical across pruning
methods, and it achieves the highest accuracy on all three models at comparable TTFT. The overhead of
DIPrune, comprising proxy scoring (attention aggregation and rank displacement) and selection (score
fusion, Top-$K$ selection, and token gathering), accounts for at most 0.74\% of TTFT and grows only
marginally with the visual token count, consistent with the linear complexity in
Appendix~\ref{supp:complexity_analysis}.

\begin{table}[H]
\centering
\small
\caption{Efficiency across architectures (88.9\% pruning ratio). TTFT (ms) and ITT (ms/token) are
measured on POPE. \textit{Overhead}: DIPrune's pruning cost, split into proxy scoring and selection, and
its ratio to TTFT. --: not reported.}
\label{tab:eff_models}
\vspace{-6pt}
\begin{tabular*}{\linewidth}{@{\extracolsep{\fill}} l ccc ccc ccc @{}}
\toprule
\multirow{2}{*}{Method} & \multicolumn{3}{c}{LLaVA-1.5-7B} & \multicolumn{3}{c}{LLaVA-NeXT-7B} & \multicolumn{3}{c}{Qwen2.5-VL-7B} \\
\cmidrule(lr){2-4}\cmidrule(lr){5-7}\cmidrule(lr){8-10}
 & TTFT$\downarrow$ & ITT$\downarrow$ & Acc$\uparrow$ & TTFT$\downarrow$ & ITT$\downarrow$ & Acc$\uparrow$ & TTFT$\downarrow$ & ITT$\downarrow$ & Acc$\uparrow$ \\
\midrule
Baseline  & 73.23 & 16.21 & 100\%  & 102.10 & 17.44 & 100\%  & 84.28 & 16.96 & 100\% \\
FastV     & 32.43 & 14.95 & 74.4\% & 43.82  & 15.43 & 87.2\% & 53.78 & 15.46 & 87.6\% \\
SparseVLM & 33.43 & 14.98 & 86.3\% & 45.25  & 15.46 & 90.6\% & 54.35 & 15.44 & --    \\
HoloV     & 31.83 & 14.94 & 93.7\% & 43.47  & 15.44 & 95.8\% & 53.52 & 15.47 & 90.5\% \\
\rowcolor{gray!15}
DIPrune   & 32.50 & 14.96 & \textbf{97.7\%} & 43.98 & 15.40 & \textbf{98.2\%} & 54.08 & 15.46 & \textbf{93.3\%} \\
\midrule
\textit{Scoring / Selection} & \multicolumn{3}{c}{0.149 / 0.091 ms} & \multicolumn{3}{c}{0.182 / 0.122 ms} & \multicolumn{3}{c}{0.163 / 0.102 ms} \\
\textit{Overhead (total)}    & \multicolumn{3}{c}{0.24 ms (0.74\%)} & \multicolumn{3}{c}{0.30 ms (0.69\%)} & \multicolumn{3}{c}{0.27 ms (0.49\%)} \\
\bottomrule
\end{tabular*}
\end{table}

\textbf{Layer-Aligned Comparison.}
\label{app:eff_align}
For Tab.~\ref{tab:efficiency}(b), we apply the selection criteria of VisionZip and HoloV at the same
pruning layer as DIPrune (L6 and L10), so that the compared methods share identical per-layer token
counts. Moving the pruning layer from L6 to L10 improves DIPrune by +3.1\%, compared with +1.5\% for
VisionZip and +1.2\% for HoloV, suggesting that DIPrune benefits more from the richer inter-layer
evolution signal at deeper layers, which corroborates the analysis in Sec.~\ref{sec:motivation}.

\textbf{End-to-End Deployment Efficiency.} We further report peak memory, total wall-clock time, and
average per-sample wall-clock time under a 66.7\% pruning ratio (Tab.~\ref{tab:eff_bench}). DIPrune
achieves the lowest memory on all benchmarks (up to 22.9\% less than SparseVLM), as its decoupled
attention computation keeps the main forward pass compatible with efficient kernels
(Appendix~\ref{supp:attn_detail}), while maintaining comparable latency and clearly higher accuracy.

\begin{table}[H]
\centering
\scriptsize
\caption{End-to-end efficiency on LLaVA-v1.5-7B (66.7\% pruning ratio). Mem: peak GPU memory;
Time: total wall-clock time; Lat.: average wall-clock time per sample.}
\label{tab:eff_bench}
\vspace{-6pt}
\setlength{\tabcolsep}{3pt}
\begin{minipage}[t]{0.325\linewidth}
\centering
\textbf{(a)} GQA\\[2pt]
\begin{tabular}{lcccc}
\toprule
Method & \makecell{Mem\\(MiB)} & \makecell{Time\\(m:s)} & \makecell{Lat.\\(ms)} & \makecell{Acc\\(\%)} \\
\midrule
SparseVLM & 19,458 & 11:56 & 56.9 & 57.6 \\
PDrop     & 15,616 & 11:56 & 56.9 & 57.1 \\
\rowcolor{gray!15}
DIPrune   & \textbf{15,174} & 11:57 & 57.0 & \textbf{61.2} \\
\bottomrule
\end{tabular}
\end{minipage}\hfill
\begin{minipage}[t]{0.325\linewidth}
\centering
\textbf{(b)} MME\\[2pt]
\begin{tabular}{lcccc}
\toprule
Method & \makecell{Mem\\(MiB)} & \makecell{Time\\(m:s)} & \makecell{Lat.\\(ms)} & \makecell{Score\\(Sum)} \\
\midrule
SparseVLM & 19,458 & 2:09 & 54.3 & 1721 \\
PDrop     & 15,616 & 2:07 & 53.5 & 1766 \\
\rowcolor{gray!15}
DIPrune   & \textbf{15,174} & \textbf{2:02} & \textbf{51.4} & \textbf{1856} \\
\bottomrule
\end{tabular}
\end{minipage}\hfill
\begin{minipage}[t]{0.325\linewidth}
\centering
\textbf{(c)} TextVQA\\[2pt]
\begin{tabular}{lcccc}
\toprule
Method & \makecell{Mem\\(MiB)} & \makecell{Time\\(m:s)} & \makecell{Lat.\\(ms)} & \makecell{Acc\\(\%)} \\
\midrule
SparseVLM & 20,170 & 7:06 & 85.2 & 56.1 \\
PDrop     & 15,980 & 7:16 & 87.2 & 56.1 \\
\rowcolor{gray!15}
DIPrune   & \textbf{15,544} & 7:10 & 86.0 & \textbf{58.5} \\
\bottomrule
\end{tabular}
\end{minipage}
\end{table}

\subsection{More Sparsification Visualization}
\label{app:visualization}
\begin{figure*}[p]
    \centering
    
    \begin{minipage}{0.495\linewidth}
        \centering
        \includegraphics[width=\linewidth, height=1.2\textheight, keepaspectratio]{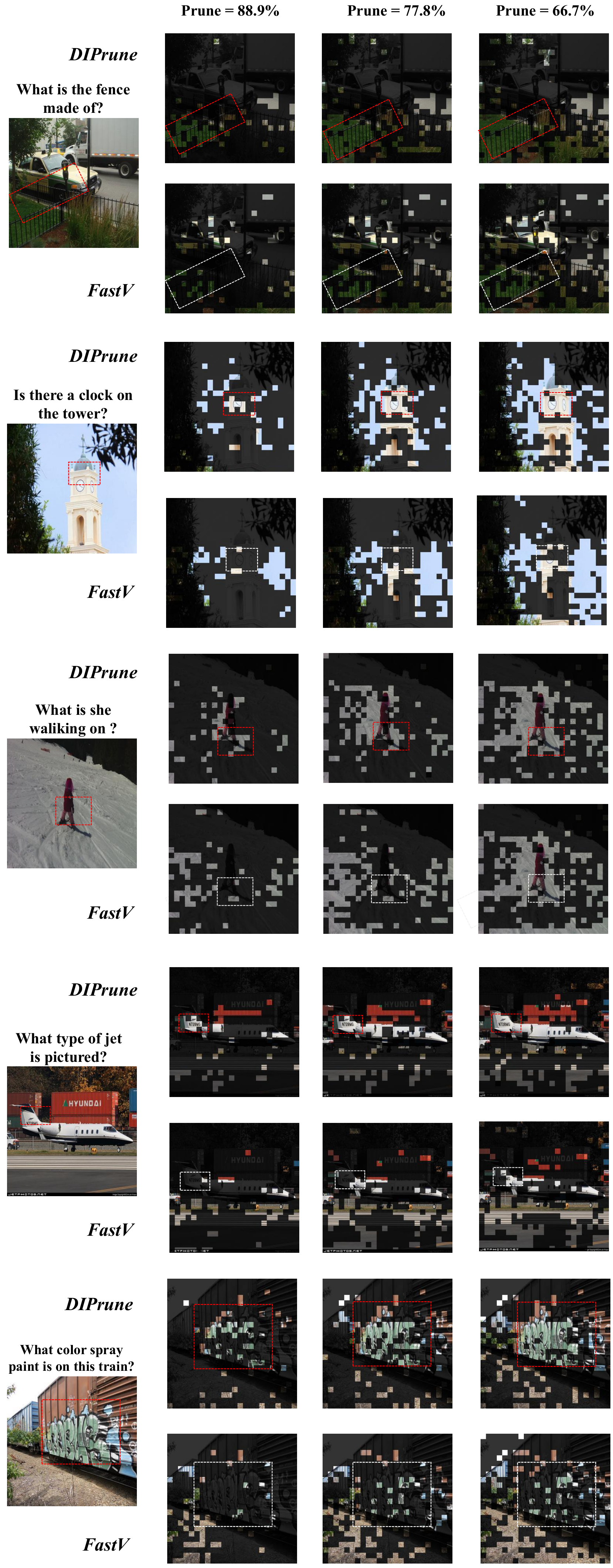}
        \centerline{\small (a)}
    \end{minipage}%
    \hspace{0.01\linewidth}%
    \begin{minipage}{0.495\linewidth}
        \centering
        \includegraphics[width=\linewidth, height=0.85\textheight, keepaspectratio]{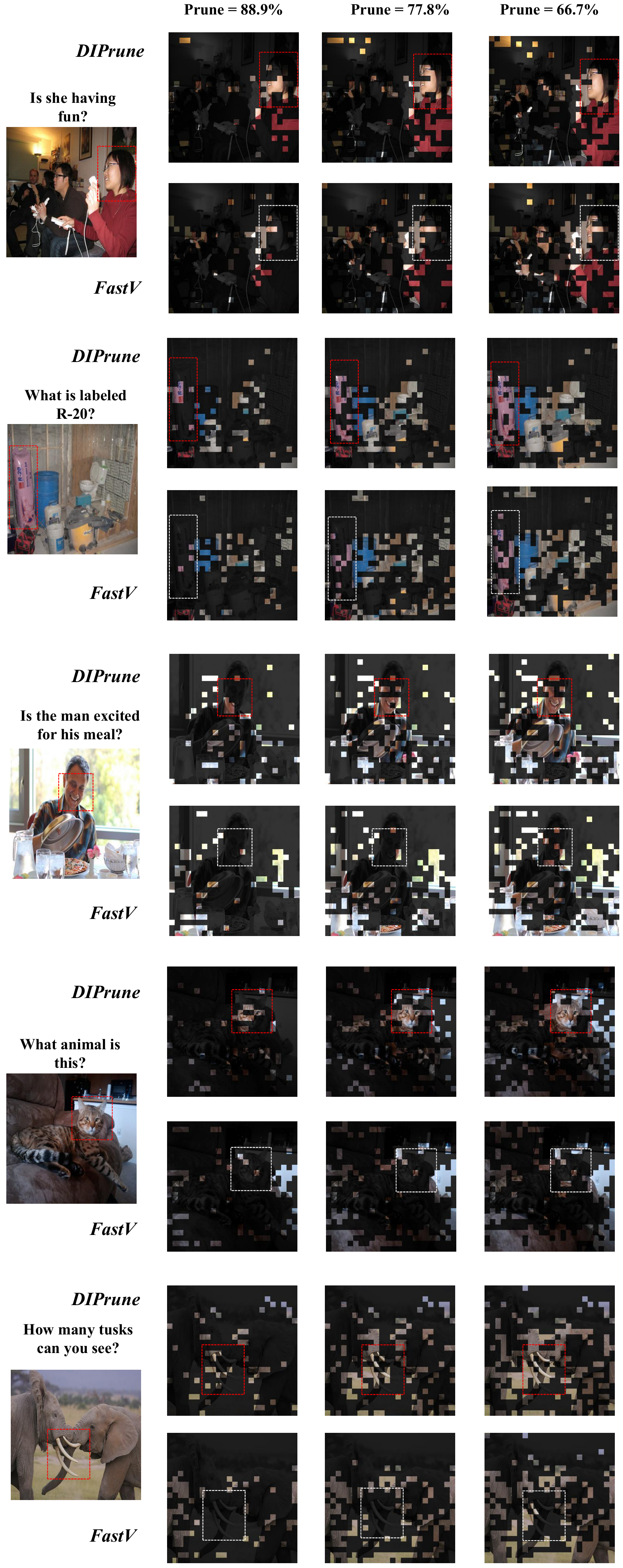}
        \centerline{\small (b)}
    \end{minipage}

    \vspace{2mm}
    
    \caption{Visualization of visual tokens retained by LLaVA-v1.5-7B on VQAv2 samples. We present comparison results under three progressive pruning ratios (66.7\%, 77.8\%, and 88.9\%). The red bounding boxes in the original images highlight the task-critical visual regions directly relevant to the text query. In the pruning results, the red and white dashed boxes indicate the retention status of these key regions by the DIPrune and FastV strategies, respectively.}    \label{fig:vis_side_by_side}
\end{figure*}

We provide detailed visualizations of the retained visual patches under varying pruning ratios (66.7\%, 77.8\%, and 88.9\%) to intuitively demonstrate the effectiveness of DIPrune. As illustrated in Figure~\ref{fig:vis_side_by_side}, the black regions denote discarded visual tokens, whereas the colored regions highlight key semantic areas that are highly aligned with the text description, showcasing how DIPrune strategically retains content with high information density. Compared to FastV, a representative pruning method based on static attention, DIPrune significantly outperforms it in preserving relevant visual cues, even under extremely high pruning ratios (e.g., 88.9\%). This superiority is achieved through our dual importance scoring mechanism. Rather than relying on isolated static attention scores, this strategy comprehensively integrates task-aware semantic alignment with dynamic inter-layer evolution. By jointly optimizing Semantic Evolution ($S_{evol}$) and Semantic Alignment ($S_{align}$), DIPrune effectively safeguards the critical features required for deep reasoning while filtering out redundant tokens.

\begin{figure*}[t!] 
    \centering
    \vspace{-28pt} 
    \includegraphics[width=1.0\textwidth]{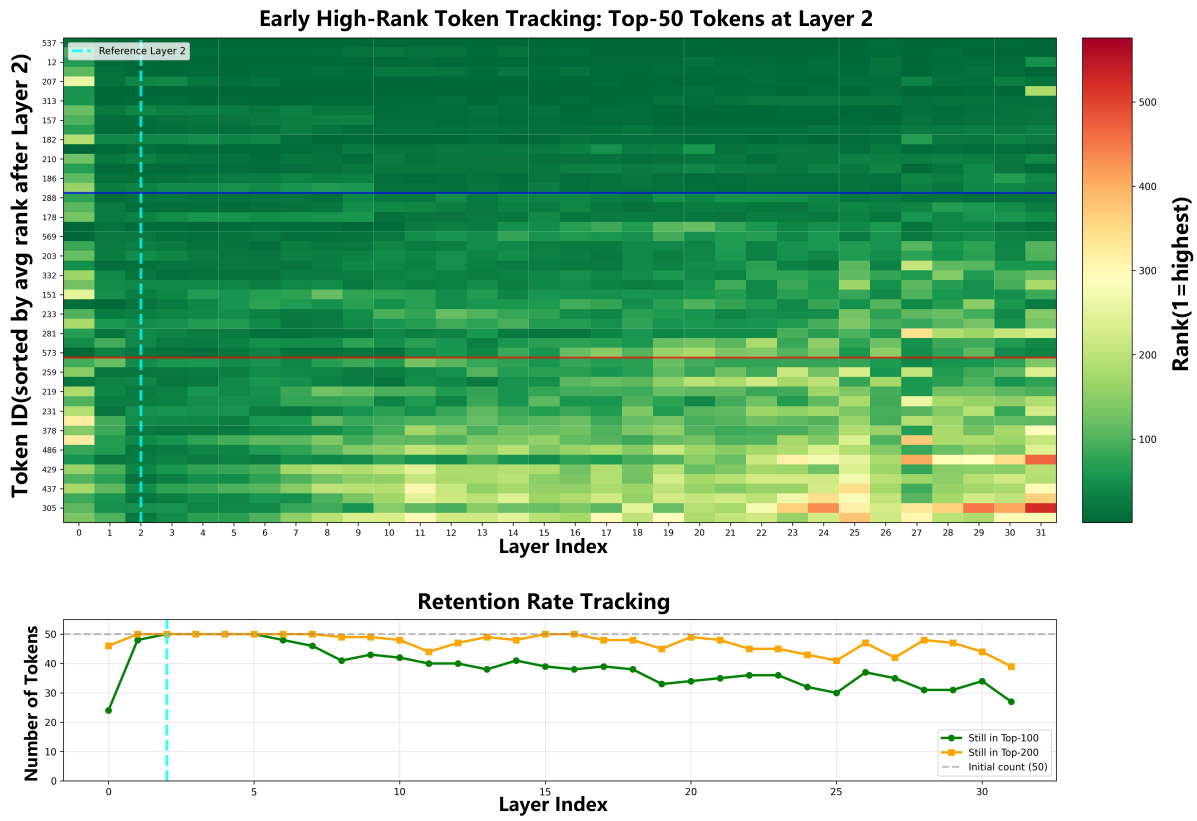} 
    
    \caption{Analysis of Numerical Inertia. Visualization of the attention score and rank evolution of the Top-50 tokens from the second layer. The heatmap shows that high-ranking tokens (green regions) in shallow layers tend to maintain their dominance throughout the network.}
    
    \label{fig:analyse_8} 
    \vspace{340pt} 
\end{figure*}

\newpage

\end{document}